\documentclass[journal]{IEEEtran}

\usepackage{url}
\usepackage{graphicx}
\usepackage{amsmath}
\usepackage{amssymb}
\usepackage[caption=false]{subfig}
\usepackage{hyperref}
\usepackage{color, colortbl}
\usepackage{mathtools}
\usepackage{array}
\usepackage{multirow}
\usepackage{subfloat}
\usepackage{pifont}
\newcommand{\cmark}{\ding{51}}%
\newcommand{\xmark}{\ding{55}}%

\newcommand{\ie}{\emph{i.e.}~}
\def\etal{\emph{et al.}~}

\usepackage{color,soul}
\usepackage[colorinlistoftodos]{todonotes}
\usepackage{xspace}
\usepackage{enumitem}

\begin{document}

\title{Predicting Human Eye Fixations via an LSTM-based Saliency Attentive Model}

\author{Marcella~Cornia,
        Lorenzo~Baraldi,
        Giuseppe~Serra,
        and~Rita~Cucchiara~\thanks{M. Cornia, L. Baraldi and R. Cucchiara are with the Department of Engineering ``Enzo Ferrari'', University of Modena and Reggio Emilia, Modena, Italy (e-mail: \{marcella.cornia, lorenzo.baraldi, rita.cucchiara\}@unimore.it).}~\thanks{G. Serra is with the Department of Computer Science, Mathematics and Physics, University of Udine, Udine, Italy (e-mail: giuseppe.serra@uniud.it).}}
\maketitle

\begin{abstract} 
Data-driven saliency has recently gained a lot of attention thanks to the use of Convolutional Neural Networks for predicting gaze fixations. In this paper we go beyond standard approaches to saliency prediction, in which gaze maps are computed with a feed-forward network, and present a novel model which can predict accurate saliency maps by incorporating neural attentive mechanisms. The core of our solution is a Convolutional LSTM that focuses on the most salient regions of the input image to iteratively refine the predicted saliency map. Additionally, to tackle the center bias typical of human eye fixations, our model can learn a set of prior maps generated with Gaussian functions. We show, through an extensive evaluation, that the proposed architecture outperforms the current state of the art on public saliency prediction datasets. We further study the contribution of each key component to demonstrate their robustness on different scenarios.
\end{abstract}



\section{Introduction}
\IEEEPARstart{V}{isual} cognition science has shown that humans, when observing a scene without a specific task to perform, do not focus on each region of the image with the same intensity. Instead, attentive mechanisms guide their gazes on salient and relevant parts~\cite{Rensink2000}. An intensive research effort has tried to emulate
such selective visual mechanisms, as computational saliency can be applied to a wide range of applications like image retargeting~\cite{Setlur2005}, object recognition~\cite{walther2002attentional}, video compression~\cite{Hadizadeh14}, tracking~\cite{Mahadevan13} and other data-dependent tasks such as image captioning~\cite{cornia2018paying}. 

Traditional saliency prediction methods have followed biological evidence by defining features that capture low-level cues such as color, contrast and texture or semantic concepts such as faces, people and text~\cite{harel2006graph,goferman2012context,judd2009learning,zhang2013saliency}. 
However, these techniques have failed to capture the wide variety of causes that contribute to defining visual saliency maps.

With the advent of deep neural networks, saliency prediction has achieved strong improvements both thanks to specific architectures and to large annotated datasets~\cite{kummerer2014deep,huang2015salicon,jetley2016end,mlnet2016}. Although these approaches went beyond the limitations of hand-crafted models, no one has yet investigated the incorporation of machine attention models~\cite{icml2015xuc15,gregor2015draw,vaswani2017attention} in saliency prediction.

Machine attention~\cite{icml2015xuc15} is a computational paradigm which sequentially attends to different parts of an input. This is usually achieved by exploiting a recurrent neural network, and by defining a compatibility measure between its internal state and regions of the input.
This paradigm has been successfully applied to image captioning~\cite{icml2015xuc15} and machine translation~\cite{chorowski15} to selectively focus on different parts of a sentence, and to action recognition~\cite{li2016videolstm} to focus on the relevant parts of a spatio-temporal volume. We argue that machine attention can also be effective for saliency prediction, as a powerful way to process saliency-specific features and to obtain an enhanced prediction.

\begin{figure}[tb]
\centering
\setlength\tabcolsep{1pt}
\begin{tabular}{cccc}
 \small{Image} & \small{Fixations} & \small{Groundtruth} & \small{Ours}  \\
\includegraphics[width=0.22\columnwidth]{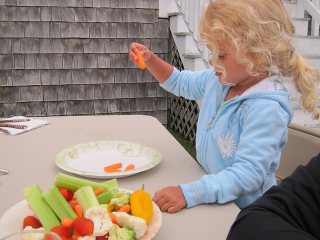}&
\includegraphics[width=0.22\columnwidth]{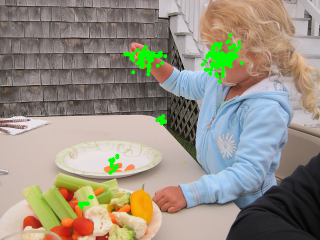}&
\includegraphics[width=0.22\columnwidth]{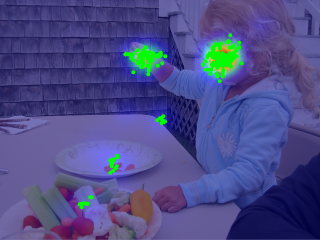}&
\includegraphics[width=0.22\columnwidth]{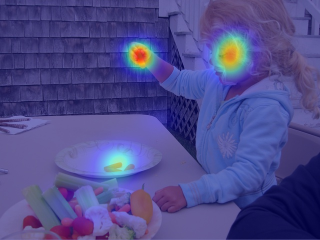}\\
\includegraphics[width=0.22\columnwidth]{images/003_img.png}&
\includegraphics[width=0.22\columnwidth]{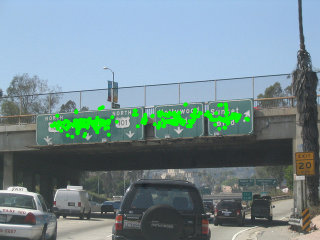}&
\includegraphics[width=0.22\columnwidth]{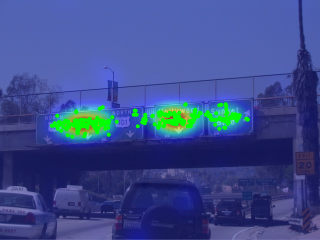}&
\includegraphics[width=0.22\columnwidth]{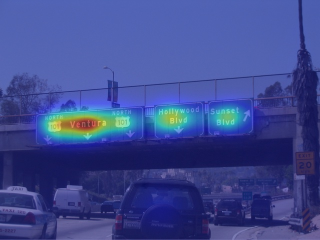}\\
\includegraphics[width=0.22\columnwidth]{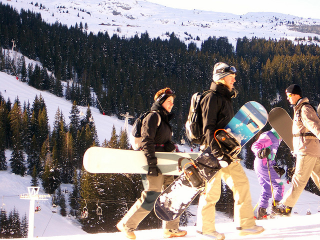}&
\includegraphics[width=0.22\columnwidth]{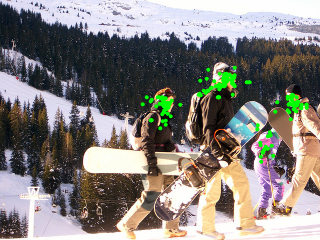}&
\includegraphics[width=0.22\columnwidth]{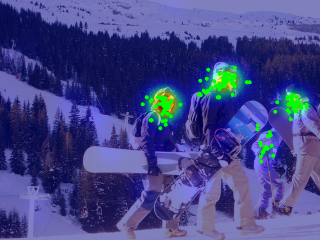}&
\includegraphics[width=0.22\columnwidth]{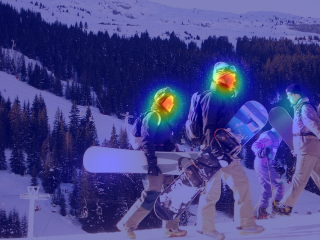}\\
\includegraphics[width=0.22\columnwidth]{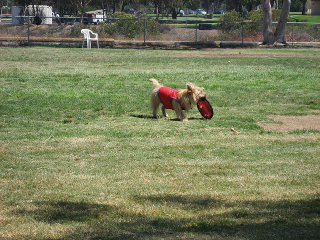}&
\includegraphics[width=0.22\columnwidth]{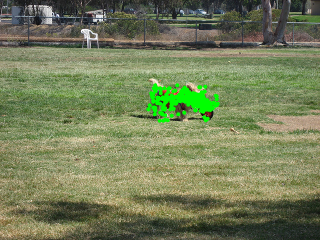}&
\includegraphics[width=0.22\columnwidth]{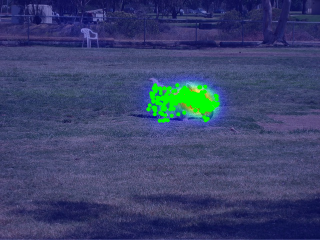}&
\includegraphics[width=0.22\columnwidth]{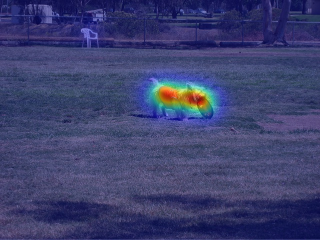}\\
\end{tabular}
\caption{Visual saliency prediction aims at predicting where humans gazes will focus on a given image. Groundtruth data is collected by means of eye-tracking glasses or mouse clicks to get eye fixation points, which are then smoothed together to obtain the groundtruth saliency map. Our model learns to predict the distribution of human fixation points by refining feature extracted from a CNN with a novel LSTM-based attentive model.}
\label{fig:first}
\end{figure}

In this paper we propose a novel saliency prediction architecture that incorporates an Attentive Convolutional Long Short-Term Memory network (Attentive ConvLSTM) that iteratively focuses on relevant spatial locations to refine saliency features. The architecture is particularly original since the LSTM model is used to achieve a refinement over an image, instead of handling a temporal sequence.

Moreover, the rescaling caused by max-pooling and strides in convolutional layers deteriorates the performance of saliency prediction, we present an extension of two popular CNNs (namely, VGG-16~\cite{Simonyan14c} and ResNet-50~\cite{he2015deep}) which can reduce the downscaling effect and maintain spatial resolution. This expedient allows us to preserve detailed visual information and obtain improved feature extraction capabilities. 

\begin{table*}[t]
\renewcommand{\arraystretch}{1.5}
\centering
\caption{Comparison between the main properties of our model and those of other existing saliency methods.}
\label{tab:related}
\begin{small}
\begin{tabular}{|p{4.4cm}|m{2.8cm}|m{2.5cm}|m{2.7cm}|m{3.1cm}|}
\hline 

& \centering CNN & \centering Attentive LSTM & \centering Center Bias & \centering Loss Function \arraybackslash \\ \hline \hline

SALICON~\cite{huang2015salicon} & \centering AlexNet - VGG-16 - GoogleNet & \centering \xmark & \centering \xmark & \centering KL-Div \arraybackslash \\ \hline

DeepFix~\cite{kruthiventi2015deepfix} & \centering VGG-16 & \centering \xmark & \centering handcrafted priors & \centering Euclidean loss \arraybackslash \\ \hline

SalNet~\cite{cPan} & \centering VGG-16 & \centering \xmark & \centering \xmark & Euclidean loss \centering \arraybackslash \\ \hline

PDP~\cite{jetley2016end} & \centering VGGNet & \centering \xmark & \centering \xmark &  probability distances \centering \arraybackslash \\ \hline

ML-Net~\cite{mlnet2016} & \centering VGG-16 & \centering \xmark & \centering single multiplicative map & \centering normalized MSE \arraybackslash \\  \hline

DSCLRCN~\cite{liu2016deep} & \centering VGG-16 - ResNet-50 & \centering \xmark & \centering \xmark & \centering NSS \arraybackslash \\ \hline

\textbf{Saliency Attentive Model (SAM)} & \centering VGG-16 - ResNet-50 & \centering \cmark & \centering multiple learned priors & \centering combination of multiple saliency metrics \arraybackslash \\ 
\hline
\end{tabular}
\end{small}
\end{table*}

Finally, in order to handle the tendency of humans to fix the center region of an image, we also introduce an explicit prior component. 
Unlike previous approaches that include handcrafted priors~\cite{judd2009learning,vig2014large,kummerer2014deep,kruthiventi2015deepfix,kummerer2016deepgaze}, our module keeps the architecture trainable end-to-end and can learn priors in an automatic way. 

Figure~\ref{fig:first} shows examples of saliency maps predicted by the proposed solution, which we call Saliency Attentive Model (SAM), compared with groundtruth saliency maps obtained from human eye fixations.
We quantitatively validate our approach on three publicly available benchmark datasets: SALICON, MIT300 and CAT2000. Experimental results will show that the proposed solution significantly improves predictions. 
\\
To summarize, the contributions of this paper are threefold: 
\begin{itemize}
\item We propose a novel Attentive ConvLSTM that sequentially focuses on different spatial locations of a stack of features to enhance predictions. To the best of our knowledge, we are among the first to incorporate attentive models in a saliency prediction architecture.
\item Our network is able to learn the bias present in eye fixations, without the need to integrate this information manually.
\item The proposed solution overcomes by a big margin the current state of the art on the largest dataset available for saliency prediction, SALICON. Moreover, on MIT300 and CAT2000 our method achieves state of the art results showing competitive generalization properties.
\end{itemize}
We make the source code of our method and pre-trained models publicly available\footnote{\url{https://github.com/marcellacornia/sam}}.

\section{Related Work}
\begin{figure*}[t!]
\centering
\includegraphics[width=1.0\textwidth]{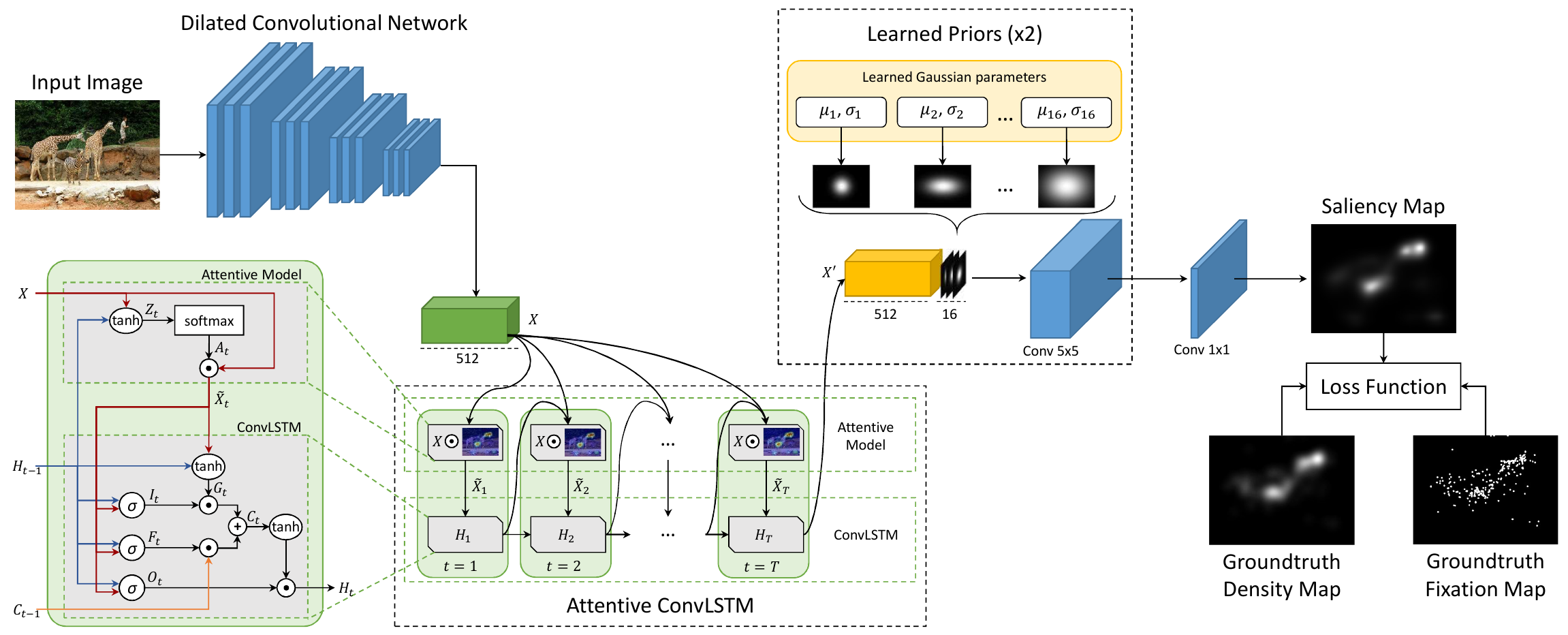} 
\caption{Overview of our Saliency Attentive Model (SAM). After computing a set of feature maps on the input image through a new architecture called Dilated Convolutional Network, an Attentive Convolutional LSTM sequentially enhances saliency features thanks to an attentive recurrent mechanism. Predictions are then combined with multiple learned priors to model the tendency of humans to fix the center region of the image. During the training phase, we encourage the network to minimize a combination of different loss functions, thus taking into account different quality aspects that predictions should meet.}
\label{fig:model}
\end{figure*}

Pioneering works on saliency prediction were based on the Feature Integration Theory proposed by Treisman~\etal\cite{treisman1980feature} in the eighties. Itti~\etal\cite{itti1998model} defined the first computational model to predict saliency on images: this work, inspired by Koch and Ullman~\cite{koch1987shifts}, computed a set of individual topographical maps representing low-level cues such as color, intensity and orientation and combined them into a global saliency map. After this seminal work, a large variety of methods explored the same idea of combining complementary low-level features~\cite{bruce2005saliency,harel2006graph,erdem2013visual} and often included additional center-surround cues~\cite{murray2011saliency,zhang2013saliency}. Other methods enriched predictions exploiting semantic classifiers for detecting higher level concepts such as faces, persons, cars and horizons~\cite{cerf2008predicting,judd2009learning,zhao2011learning,goferman2012context,borji2012boosting}.
Related research efforts have also been done in the compressed domain, as in~\cite{fang2012saliency,fang2014video}. 

\subsection{Saliency and Deep Learning}
Only recently, thanks to the large spread of deep learning techniques, the saliency prediction task has achieved a considerable improvement. One of the first proposals has been the \textit{Ensemble of Deep Networks} (\textit{e}DN) model by Vig~\etal\cite{vig2014large}. This model consists of three convolutional layers followed by a linear classifier that blends feature maps coming from the previous layers.
After this work, K{\"u}mmerer~\etal\cite{kummerer2014deep,kummerer2016deepgaze} proposed two deep saliency prediction networks: the first, called \textit{DeepGaze I}, was based on the AlexNet model~\cite{krizhevsky2012imagenet}, while the second, \textit{DeepGaze II}, was built upon the VGG-19 network~\cite{Simonyan14c}. Liu~\etal\cite{liu2015predicting} presented a multi-resolution CNN (\textit{Mr-CNN}) fine-tuned over image patches centered on fixation and non-fixation locations.

It is well known that deep learning approaches strongly depend on the availability of sufficiently large datasets. The publication of a large-scale eye-fixation dataset, SALICON~\cite{jiang2015salicon}, indeed contributed to a big progress of deep saliency prediction models. Huang~\etal\cite{huang2015salicon} introduced an architecture consisting of a deep neural network applied at two different image scales. They compared different standard CNN architectures such as AlexNet~\cite{krizhevsky2012imagenet}, VGG-16~\cite{Simonyan14c} and GoogleNet~\cite{szegedy2015going}, in particular showing the effectiveness of the VGG network.

After this work, several deep saliency models based on the VGG network have been published~\cite{kruthiventi2015deepfix, cPan, jetley2016end, kruthiventi2016saliency, mlnet2016, tavakoli2016exploiting, pan2017salgan, dodge2017visual}. Accordingly, we proposed a new architecture, called \textit{ML-Net}~\cite{mlnet2016}, which improved previous attempts by using features coming from multiple layers of a CNN and by adding a learned prior map. In particular, we learned a matrix of weights which was applied to the output saliency map with a pixel-wise multiplication. The usage of centered priors has also been investigated in~\cite{kruthiventi2015deepfix}, where multiple predefined priors were fed to a convolutional layer.

In this work, instead, we model the center bias present in human gazes using multiple learned prior maps. This is different from the approaches of~\cite{mlnet2016} and~\cite{kruthiventi2015deepfix}, as we let the network learn a set of Gaussian parameters, keeping it trainable end-to-end without predefined information. 

Recently, Pan~\etal\cite{pan2017salgan} introduced \textit{SalGAN}, a deep network for saliency prediction trained with adversarial examples. As all other Generative Adversarial Networks, it is composed by two modules, a generator and a discriminator, which combine efforts to produce saliency maps.

In this work, we also employ the ResNet~\cite{he2015deep} model to extract feature maps from the input image. The only other saliency model that exploits this network is proposed by Liu~\etal\cite{liu2016deep} and called \textit{DSCLRCN}. This model simultaneously incorporates global and scene contexts to infer image saliency thanks to a deep spatial contextual LSTM which scans the image both horizontally and vertically.

To better highlight the differences of our model with respect to other existing saliency methods, we report in Table~\ref{tab:related} a summary of the main properties of our solution and those of the most competitive methods. Note that none of the other methods incorporate an attentive mechanism or a set of prior maps directly learned by the network. In addition, differently from other previous models, we propose a loss function which is a balanced combination of different saliency metrics and that provides state of the art performances.

A related line of research is that of explaining activations of a neural model by means of techniques based on backpropagation~\cite{zhang2016top}. It is worthwhile to notice that this research line is very different from that of saliency prediction, as it does not aim to replicate human fixations.

\subsection{Salient Object Detection}
Salient object detection is slightly related to the topic of this work, even though it is a significantly different task. Salient object detection consists, indeed, in identifying a binary map indicating the presence of salient objects~\cite{li2015visual,wang2015deep,zhang2015co,zhao2015saliency}. On the contrary, in saliency prediction the objective is to predict a density map of eye fixations.

A saliency detection approach which is in some aspects related to our work is that of Kuen~\etal\cite{kuen2016recurrent}, in which a recurrent (non convolutional) network provides salient object detection. At each timestep, their recurrent network outputs the parameters of a spatial transformation which is used to focus on a particular location of the image, and builds the binary prediction for that location. Our recurrent network is, instead, convolutional, and is used to process saliency features by iteratively refining the prediction.

\section{Model Architecture}
In this section we present the architecture of our complete model, called SAM (Saliency Attentive Model). 

The main novelty of our proposal is an Attentive Convolutional model, which recurrently processes saliency features at different locations, by selectively attending to different regions of a tensor. This architecture, that for the first time uses an LSTM without the concept of time, is described in Section~\ref{sec:attentive_lstm}.

Predictions are then combined with multiple learned priors which are used to model the human-gaze center bias (Section~\ref{sec:learned_priors}). To extract feature maps from input images, we employ a Convolutional Neural Network model. Instead of using a pre-defined CNN, we propose a Dilated Convolutional Network to limit the rescaling effects which can worse saliency prediction performance (Section~\ref{sec:dilated_cnn}).
 A new combination of different loss functions is finally used to train the whole network by simultaneously taking into account different quality aspects (Section~\ref{sec:training}).
The overall architecture of our model is shown in Figure~\ref{fig:model}. 

\subsection{Attentive Convolutional LSTM} \label{sec:attentive_lstm}
\begin{figure}[t]
\centering
\setlength\tabcolsep{1pt}
\begin{tabular}{ccccc}
 \small{$t=1$} & \small{$t=2$} & \small{$t=3$} & \small{$t=4$} & \small{GT}  \\
\includegraphics[width=0.18\columnwidth]{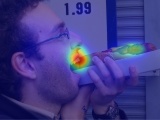}&
\includegraphics[width=0.18\columnwidth]{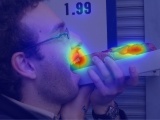}&
\includegraphics[width=0.18\columnwidth]{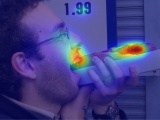}&
\includegraphics[width=0.18\columnwidth]{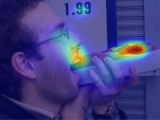}&
\includegraphics[width=0.18\columnwidth]{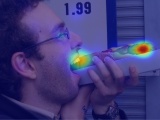}\\
\includegraphics[width=0.18\columnwidth]{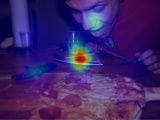}&
\includegraphics[width=0.18\columnwidth]{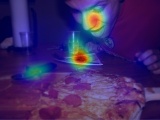}&
\includegraphics[width=0.18\columnwidth]{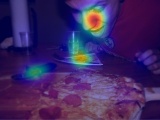}&
\includegraphics[width=0.18\columnwidth]{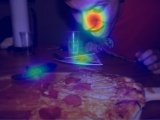}&
\includegraphics[width=0.18\columnwidth]{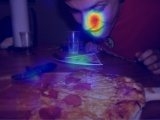}\\
\includegraphics[width=0.18\columnwidth]{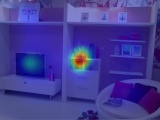}&
\includegraphics[width=0.18\columnwidth]{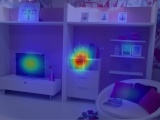}&
\includegraphics[width=0.18\columnwidth]{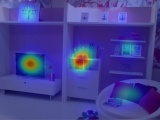}&
\includegraphics[width=0.18\columnwidth]{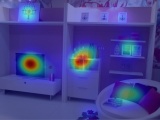}&
\includegraphics[width=0.18\columnwidth]{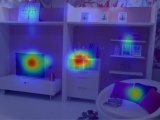}\\
\includegraphics[width=0.18\columnwidth]{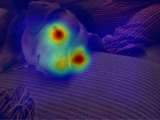}&
\includegraphics[width=0.18\columnwidth]{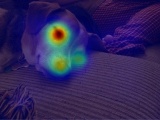}&
\includegraphics[width=0.18\columnwidth]{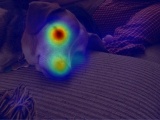}&
\includegraphics[width=0.18\columnwidth]{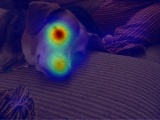}&
\includegraphics[width=0.18\columnwidth]{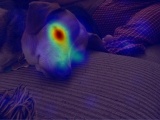}\\
\end{tabular}

\caption{Progressive refinement of predictions performed by the Attentive ConvLSTM. The first and the second row show a progressive change of focus in the saliency map, so that regions which were wrongly predicted as salient are progressively corrected, and truly salient regions are correctly identified. The third and the fourth row, instead, respectively show an increase and a reduction of saliency in regions of the image that have been (or have not been) considered as salient at the first timestep. In all cases, the result is a progressive approach of the saliency map to the groundtruth.}
\label{fig:attentive}
\end{figure}
Long Short-Term Memory networks~\cite{hochreiter1997long} achieved good performances on several tasks in which time dependencies are a key component~\cite{donahue2015long,karpathy2015deep,wu2015ask,baraldi17cvpr}, but they can not be directly employed for saliency prediction, as they work on sequences of time varying vectors. We extend the traditional LSTM to work on spatial features: formally this is achieved by substituting dot products with convolutional operations in the LSTM equations. Moreover, we exploit the sequential nature of LSTM to process features in an iterative way, instead of using the model to deal with temporal dependencies in the input.

To explain our proposal of the attentive model, let's consider the LSTM scheme on the left part of Fig.~\ref{fig:model}. Here the LSTM takes as input a stack of features extracted from the input image ($X$ in Fig.~\ref{fig:model}) and produces a refined stack of feature maps ($X'$ in Fig.~\ref{fig:model}) entering in the learned prior module. The LSTM works by sequentially updating an internal state, according to the values of three sigmoid gates. Specifically, the update is driven by the following equations:
\begin{align}
I_t &= \sigma (W_{i} * \tilde{X}_t + U_{i} * H_{t-1} + b_i) \\
F_t &= \sigma (W_{f} * \tilde{X}_t + U_{f} * H_{t-1} + b_f) \\
O_t &= \sigma (W_{o} * \tilde{X}_t + U_{o} * H_{t-1} + b_o) \\
G_t &= \tanh (W_{c} * \tilde{X}_t + U_{c} * H_{t-1} + b_c) \\
C_t &= F_t \odot C_{t-1} + I_t \odot G_t \\
H_t &= O_t \odot \tanh(C_t).
\end{align}
Here, the gates $I_t$, $F_t$, $O_t$, the candidate memory $G_t$, memory cell $C_t$, $C_{t-1}$, and hidden state $H_t$, $H_{t-1}$ are 3-d tensors, each of them having 512 channels. $*$ represents the convolutional operator, all $W$ and $U$ are 2-d convolutional kernels, and all $b$ are learned biases.

The input of the LSTM layer $\tilde{X}_t$ is computed, at each timestep (\ie at each iteration), through an attentive mechanism.
In particular, an attention map is generated by convolving the previous hidden state $H_{t-1}$ and the input $X$, feeding the result to a $\tanh$ activation function and finally convolving with a one channel convolutional kernel:
\begin{equation}
Z_t = V_a * \tanh (W_{a} * X + U_{a} * H_{t-1} + b_a).
\end{equation}
The output of this operations is a 2-d map from which we can compute a normalized spatial attention map through the \textit{softmax} operator:
\begin{equation}
A^{ij}_t = p(att_{ij}|X, H_{t-1}) = \frac{\exp(Z^{ij}_t)}{\sum_i \sum_j \exp(Z^{ij}_t)}
\end{equation}
where $A^{ij}_t$ is the element of the attention map in position $(i, j)$. The attention map is applied to the input $X$ with an element-wise product between each channel of the feature maps and the attention map:
\begin{equation}
\tilde{X}_t = A_t \odot X.
\end{equation} 

Fig.~\ref{fig:attentive} shows saliency predictions on four sample images, using the output of the ConvLSTM module at different timesteps as input of the rest of the model. As can be noticed, predictions are progressively refined by modifying the initial map given by the CNN. This refinement results in an significant enhancement of the predictions.

\begin{figure*}[t!]
\centering
\subfloat[Dilated VGG Convolutional Network] {
	\includegraphics[width=0.842\textwidth]{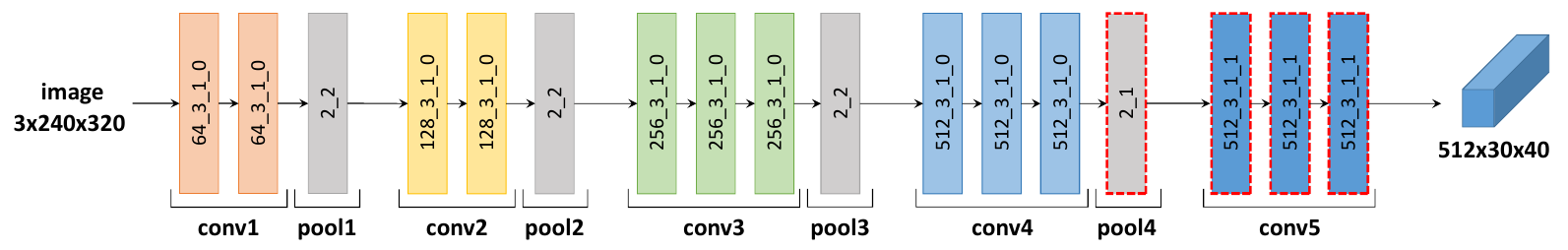}\label{fig:vgg}
} \\
\subfloat[Dilated Residual Convolutional Network] {
	\includegraphics[width=0.95\textwidth]{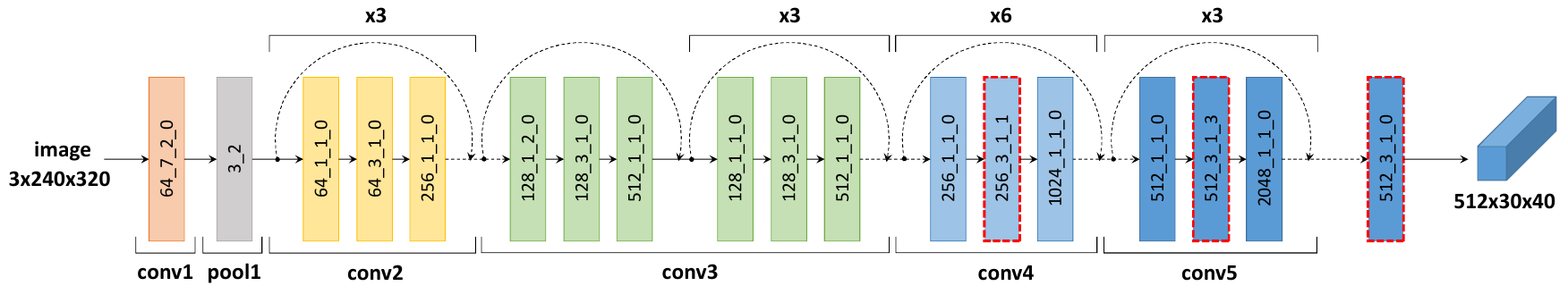}\label{fig:resnet}
}
\caption{Overall architectures of Dilated Convolutional Networks based on the VGG-16 and ResNet-50 models. Convolutional and pooling blocks are respectively expressed in terms of \texttt{channels\_kernel\_stride\_holes} and \texttt{kernel\_stride}. On top of the ResNet model, we report the number of repetitions for each block. Red dashed edges indicate modified layers with respect to the original networks.}
\label{fig:architecture}
\end{figure*}

\subsection{Learned Priors}
\label{sec:learned_priors}
Psychological studies have shown that when observers look at images, their gazes are biased toward the center~\cite{tatler2007central,tseng2009quantifying}. 
This phenomenon is mainly due to the tendency of photographers to position objects of interest at the center of the image. Also, when people repeatedly watch images with salient information placed in the center, they naturally expect to find the most informative content of the image around its center~\cite{tseng2009quantifying}. Another important reason that encourages this behavior is the interestingness of the scene~\cite{CAT2000}. Indeed, when there are no highly salient regions, humans are inclined to look at the center of the image.

Based on this evidence, the inclusion of center priors is a key component of several recent works of saliency prediction~\cite{judd2009learning,vig2014large,kummerer2014deep,kruthiventi2015deepfix,kummerer2016deepgaze,mlnet2016}. Differently from existing works, which included pre-defined priors, we let the network learn its own priors. To reduce the number of parameters and facilitate the learning, we constraint each prior to be a 2d Gaussian function, whose mean and covariance matrix are instead freely learnable. This lets the network learn its own priors purely from data, without relying on assumptions from biological studies.

We model the center bias by means of a set of Gaussian functions with diagonal covariance matrix. Means and variances are learned for each prior map according to the following equation: 
\begin{equation}
f(x, y) = \frac{1}{2 \pi \sigma_x \sigma_y} \exp \left(- \left(\frac{(x - \mu_x)^2}{2\sigma^2_x} + \frac{(y - \mu_y)^2}{2\sigma^2_y} \right) \right).
\end{equation}

Our network learns the parameters of $N$ Gaussian functions (in our experiments $N=16$) and generates the relative prior maps. Since the $X'$ tensor has $512$ channels, after the concatenation with learned prior maps, we obtain a tensor with $528$ channels. The resulting tensor is fed through a convolutional layer with $512$ filters. This operation adds more non-linearity to the model and proves to be effective with respect to other previous works, as reported in Section~\ref{sec:exp_priors}. The entire prior learning module is replicated two times.

\subsection{Dilated Convolutional Network}
\label{sec:dilated_cnn}
One of the main drawbacks of using CNNs to extract features for saliency prediction is that they considerably rescale the input image during the feature extraction phase, thus worsening the prediction accuracy. In the following, we devise a strategy which increases the output resolution of a CNN while preserving the scale at which convolutional filters operate and the number of parameters. This makes it possible to use pre-trained weights, and thus to reduce the need for fine-tuning convolutional filters after the network structure has been modified.

The intuition of the approach is that given a CNN of choice and one of its layers having stride $s > 1$, we can increase the output resolution by reducing the stride of the layer, and adding dilation~\cite{yu2015multi} to all the layers which follow the chosen layer. In this way, all convolutional filters still operate on the same scale they have been trained for.
We apply this technique on two recent feature extraction networks: the VGG-16~\cite{Simonyan14c} and the ResNet-50~\cite{he2015deep}.

The VGG-16 network is composed by 13 convolutional layers and 3 fully connected layers. The convolutional layers are divided in five convolutional blocks where, each of them is followed by a max-pooling layer with a stride of 2.

The ResNet-50, instead of having a series of stacked layers that process the input image as in common CNNs, performs a series of residual mappings between blocks composed by a few stacked layers. This is obtained using shortcut connections that realize an identity mapping, \ie the input of the block is added to its output. Residual connections help to avoid the accuracy degradation problem~\cite{he2015convolutional} that occurs with the increase of the network depth, and are beneficial also in the saliency prediction case, since they improve the feature extraction capabilities of the network. 

In particular, the ResNet-50 network consists of five convolutional blocks and a fully connected layer. The first block is composed by one convolutional layer followed by a max-pooling layer, both of them having a stride of 2, while the remaining four blocks are fully convolutional. All of these blocks, except the second one (\texttt{conv2}), reduce the dimension of feature maps with strides of 2. 

Since the purpose of our network is to extract feature maps, we only consider convolutional layers and ignore fully connected layers which are present at the end of both networks. Moreover, it can be noticed that the downscaling factor of both of these architectures is particularly critical. For example, with an input image having a size of $240 \times 320$, the output dimension is $8 \times 10$, which is relatively small for the saliency prediction task. For this reason, we modify network structures to limit the rescaling phenomenon. 

For the VGG-16 model, we also remove the last max-pooling layer and apply the aforementioned technique to the last but one pooling layer (see Figure~\ref{fig:vgg}).
On the contrary, for the ResNet-50 model we remove the stride and we introduce dilated convolutions in the last two blocks (see Figure~\ref{fig:resnet}). In this case, since the technique is applied two times, we introduce holes of size 1 in the kernels of the block \texttt{conv4} and holes of size $2^2-1=3$ in the kernels of the block \texttt{conv5}. The output of the residual network is a tensor with 2048 channels. To limit the number of feature maps, we feed this tensor into another convolutional layer with 512 filters.
Thanks to these expedients, our saliency maps are rescaled by a factor of 8 instead of 32 as in the original VGG-16 and ResNet-50 models. 

We include dilated convolutions also in prior layers, thus obtaining two convolutional layers with large receptive fields that allow us to capture the saliency of an object with respect to its neighborhood. We set the kernel size of these layers to 5 and the holes size to 3 achieving therefore a receptive field of $17 \times 17$. Strides of these layers are set to $1$ and both of them are followed by a ReLU activation function.

The last layer of our model is a convolutional operation with one filter and a kernel size of 1 that extracts the final saliency map. Since the predicted map has lower dimensions than the original image, it is brought to its original size via bilinear upsampling.

\subsection{Loss function}
\label{sec:training}
In order to capture several quality factors, saliency predictions are usually evaluated through different metrics. Inspired by this evaluation protocol, we introduce a new loss function given by a linear combination of three different saliency evaluation metrics. We define the overall loss function as follows:
\begin{equation}
\begin{split}
L(\mathbf{\tilde{y}}, \mathbf{y}^{den}&, \mathbf{y}^{fix}) = \\ &\alpha L_1(\mathbf{\tilde{y}}, \mathbf{y}^{fix}) + \beta L_2(\mathbf{\tilde{y}}, \mathbf{y}^{den}) + \gamma L_3(\mathbf{\tilde{y}}, \mathbf{y}^{den})
\end{split}
\end{equation}
where $\mathbf{\tilde{y}}$, $\mathbf{y}^{den}$ and $\mathbf{y}^{fix}$ are respectively the predicted saliency map, the groundtruth density distribution and the groundtruth binary fixation map, while $\alpha$, $\beta$ and $\gamma$ are three scalars which balance the three loss functions. $L_1$, $L_2$ and $L_3$ are respectively the Normalized Scanpath Saliency (NSS), the Linear Correlation Coefficient (CC) and the Kullback-Leibler Divergence (KL-Div) which are commonly used to evaluate saliency prediction models.

The NSS metric was defined specifically for the evaluation of saliency models~\cite{peters2005components}. The idea is to quantify the saliency map values at the eye fixation locations and to normalize it with the saliency map variance:
\begin{equation}
L_1\left(\mathbf{\tilde{y}}, \mathbf{y}^{fix}\right) = \frac{1}{N} \sum_i \frac{\mathbf{\tilde{y}}_i - \mu(\mathbf{\tilde{y}})}{\sigma(\mathbf{\tilde{y}})} \cdot \mathbf{y}_i^{fix}
\end{equation}
where $i$ indexes the $i^{th}$ pixel, $N = \sum_i \mathbf{y}_i^{fix}$ is the total number of fixated pixels and $\mathbf{\tilde{y}}$ is normalized to have a zero mean and unit standard deviation.

The CC, instead, is the Pearson's correlation coefficient and treats the saliency and groundtruth density maps, $\mathbf{\tilde{y}}$ and $\mathbf{y}_{den}$ , as random variables measuring the linear relationship between them. It is computed as:
\begin{equation}
L_2\left(\mathbf{\tilde{y}}, \mathbf{y}^{den}\right) = \frac{\sigma\left(\mathbf{\tilde{y}}, \mathbf{y}^{den}\right)}{\sigma\left(\mathbf{\tilde{y}}\right) \cdot \sigma\left(\mathbf{y}^{den}\right)}
\end{equation}
where $\sigma\left(\mathbf{\tilde{y}}, \mathbf{y}^{den}\right)$ is the covariance of $\mathbf{\tilde{y}}$ and $\mathbf{y}^{den}$. 

The KL-Div evaluates the loss of information when the distribution $\mathbf{\tilde{y}}$ is used to approximate the distribution $\mathbf{y}^{den}$, therefore taking a probabilistic interpretation of saliency and groundtruth density maps. Formally:
\begin{equation}
L_3\left(\mathbf{\tilde{y}}, \mathbf{y}^{den}\right) = \sum_i \mathbf{y}_i^{den} \log \left( \frac{\mathbf{y}_i^{den}}{\mathbf{\tilde{y}}_i + \epsilon} + \epsilon \right)
\end{equation}
where $i$ indexes the $i^{th}$ pixel and $\epsilon$ is a regularization constant. The KL-Div is a dissimilarity metric and a lower value indicates a better approximation of the groundtruth by the predicted saliency map. 

In Section~\ref{sec:loss}, we quantitatively justify the choice of our loss combination comparing our results with those obtained using single evaluation metrics as loss function. Moreover, we compare the proposed training strategy with several other probability distances used by previous saliency methods demonstrating that our solution is able to achieve a better balance among all evaluation metrics.

\section{Experimental Setup}
In this section we describe datasets and metrics used to evaluate the proposed model, and provide implementation details.

\begin{figure*}[t!]
\centering
\small{SALICON} \\
\subfloat[CC] {
	\includegraphics[width=0.475\columnwidth]{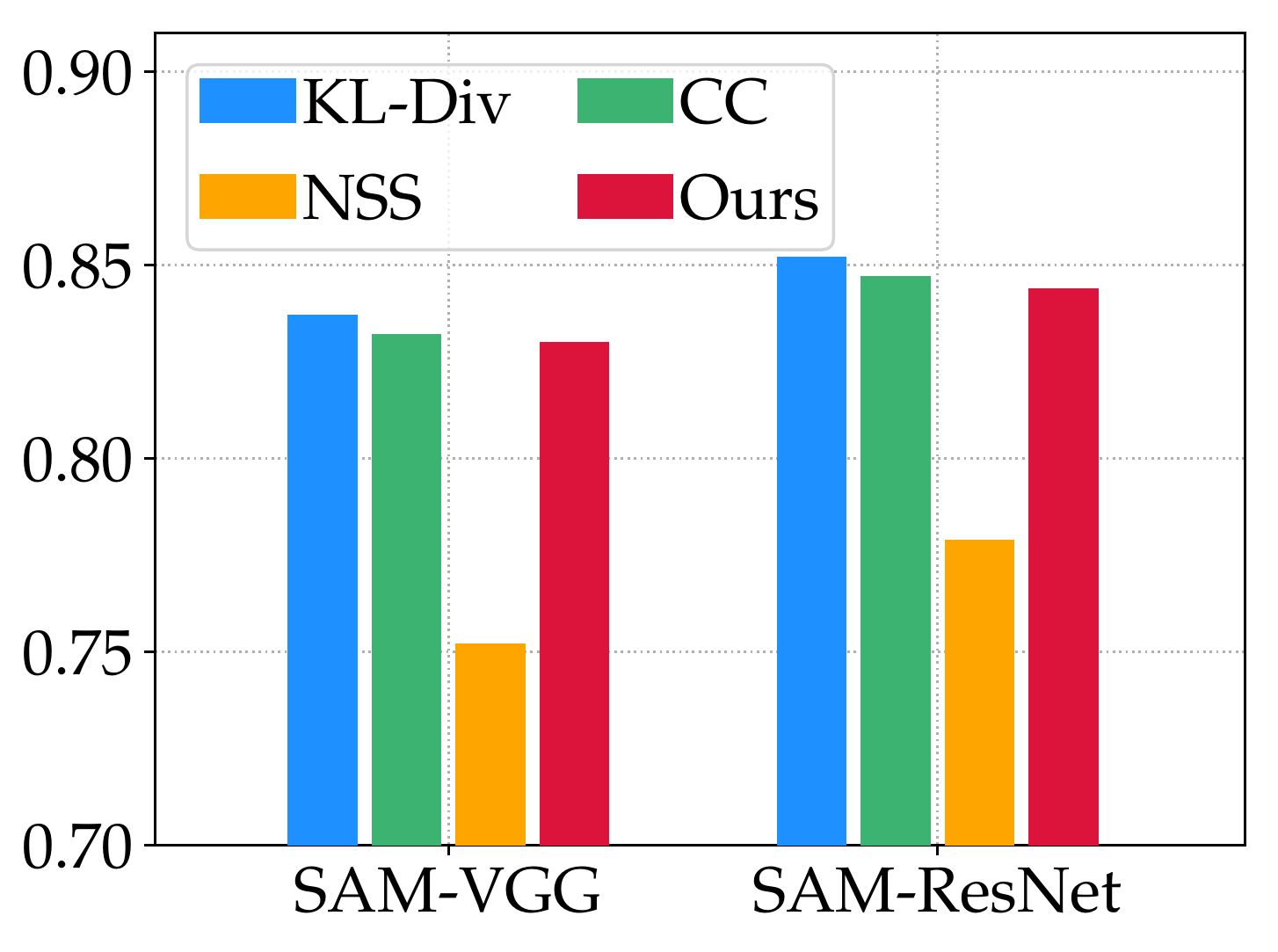}
}
\subfloat[sAUC] {
	\includegraphics[width=0.475\columnwidth]{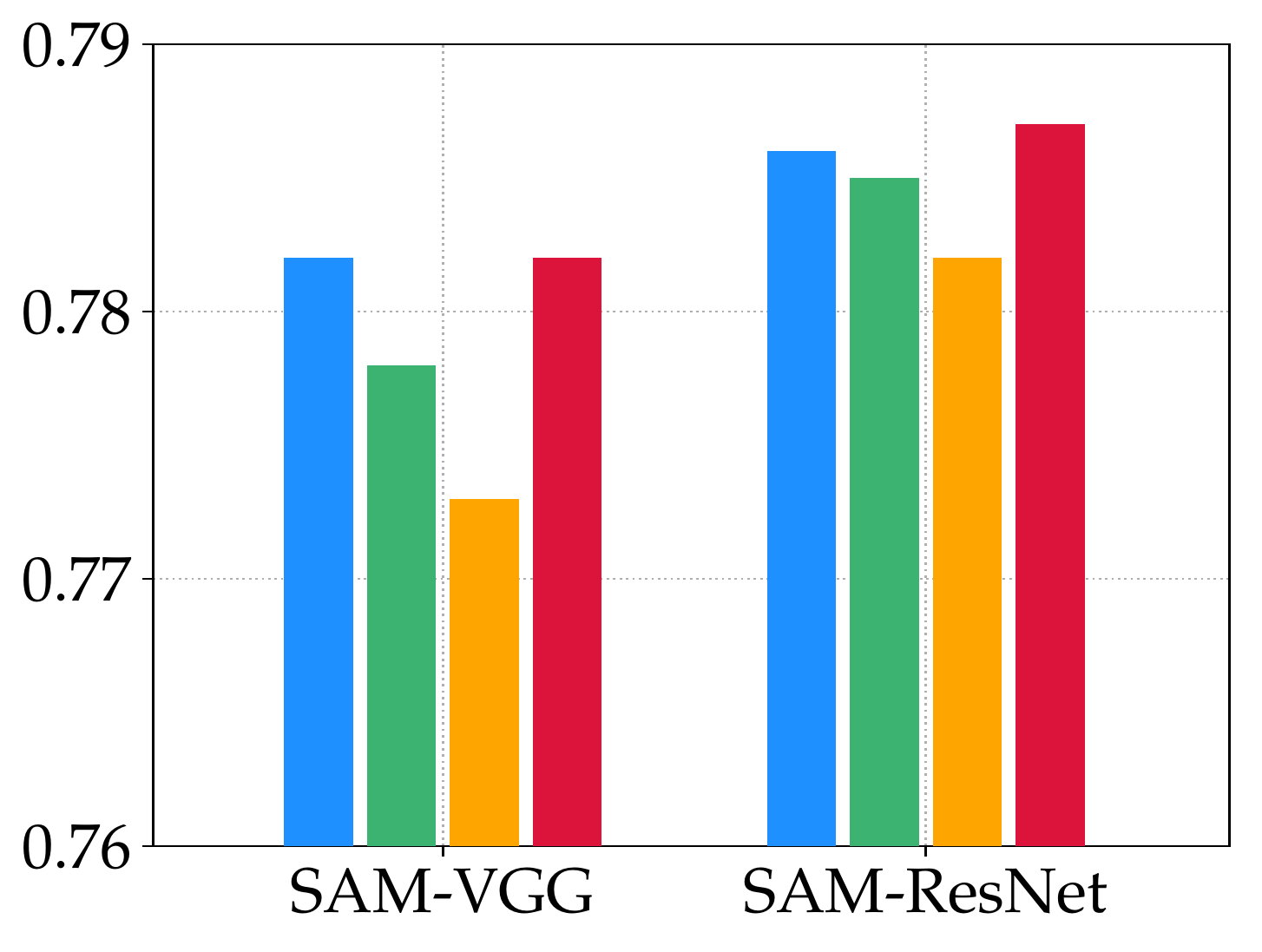}
}
\subfloat[AUC] {
	\includegraphics[width=0.475\columnwidth]{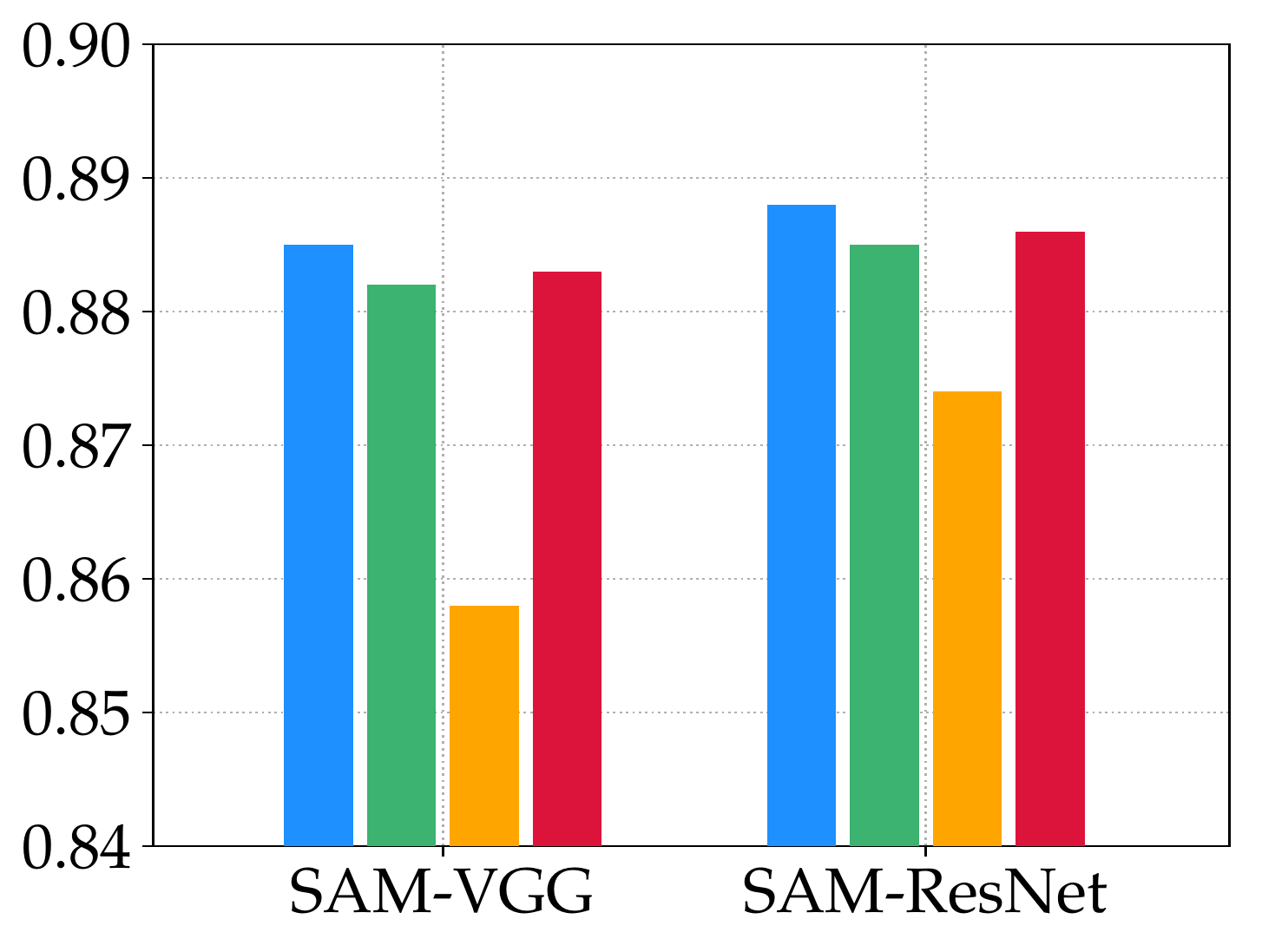}
}
\subfloat[NSS] {
	\includegraphics[width=0.475\columnwidth]{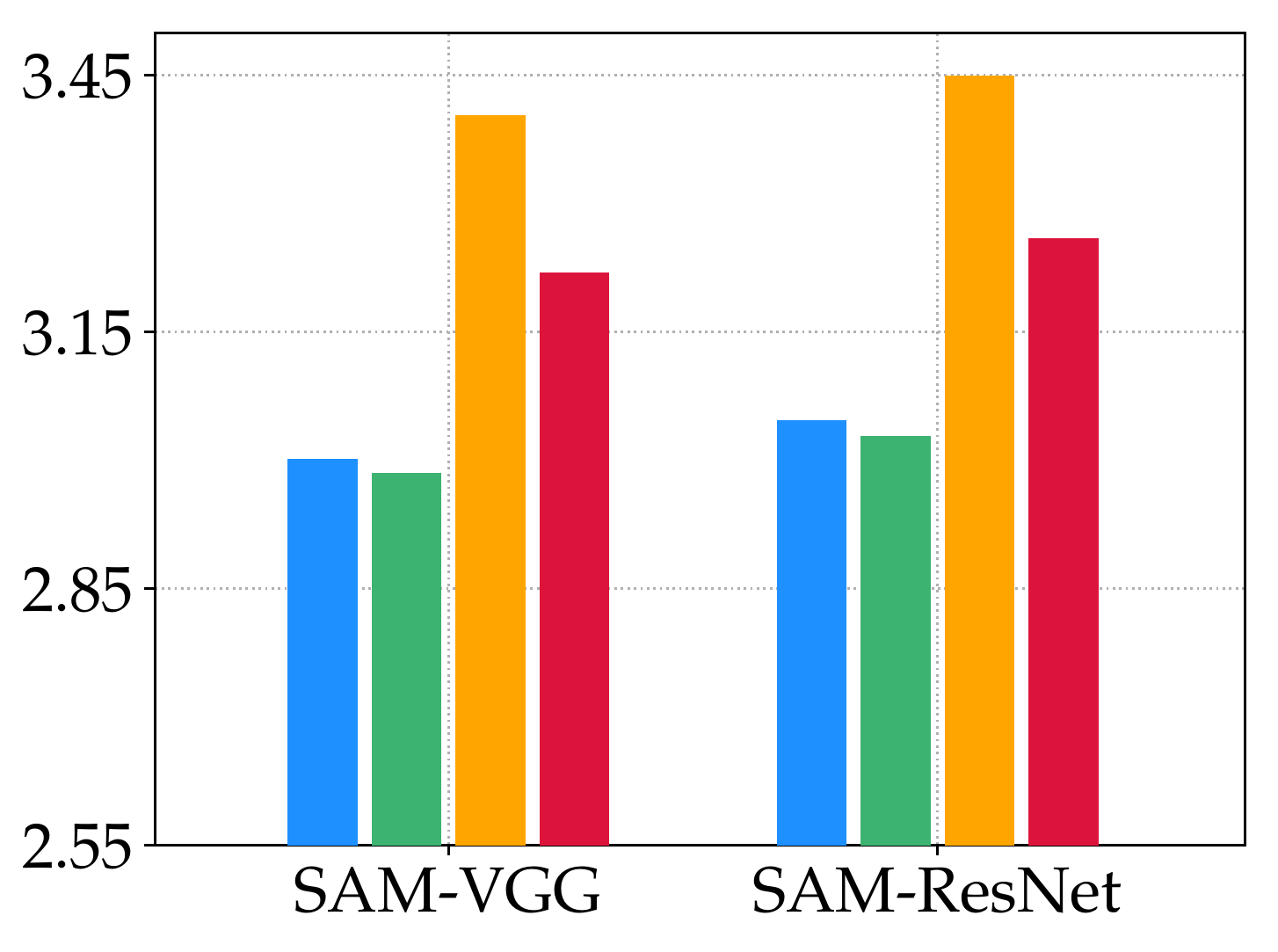}
}
\\
\small{MIT1003} \\
\subfloat[CC] {
	\includegraphics[width=0.475\columnwidth]{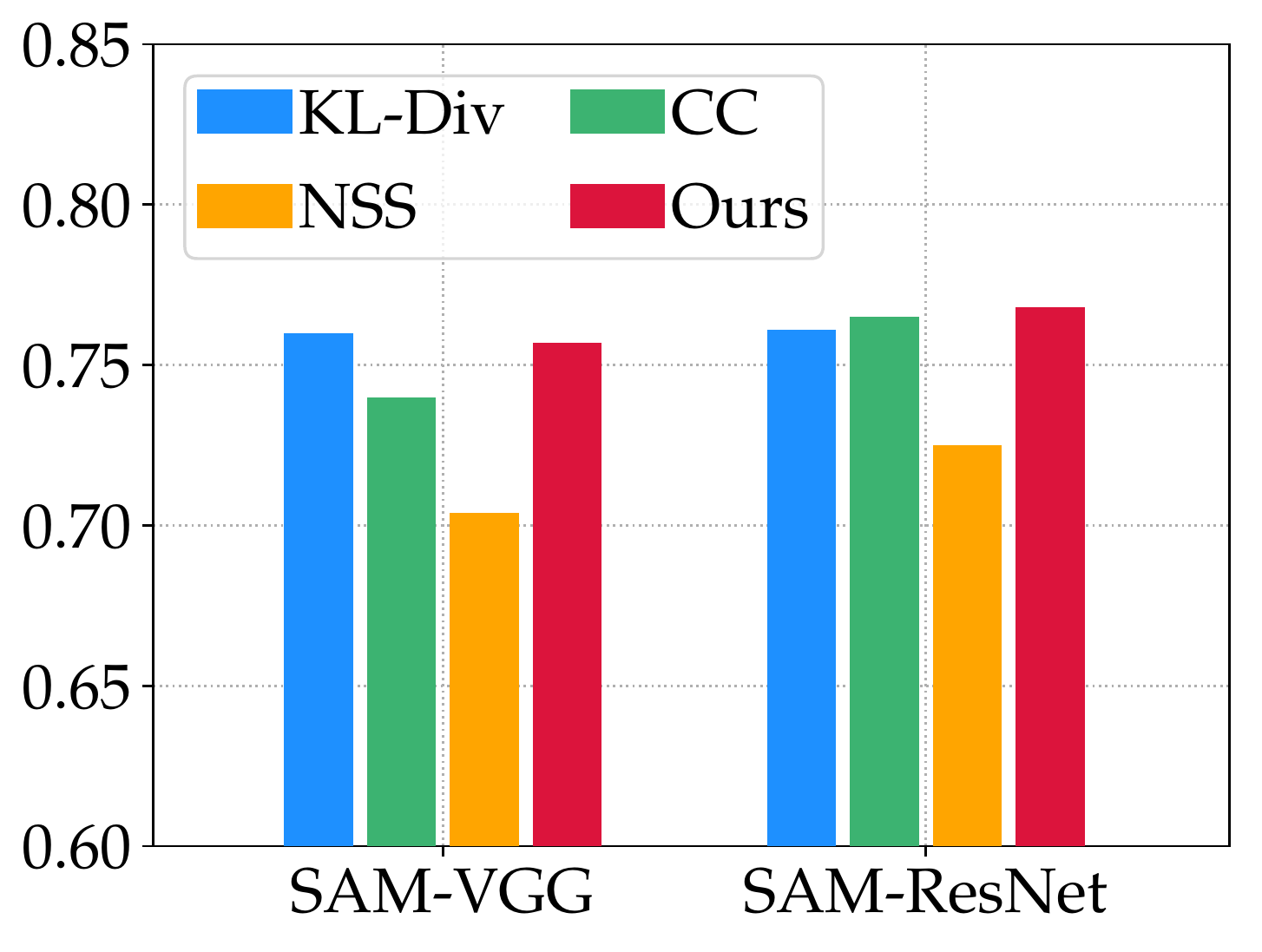}
}
\subfloat[sAUC] {
	\includegraphics[width=0.475\columnwidth]{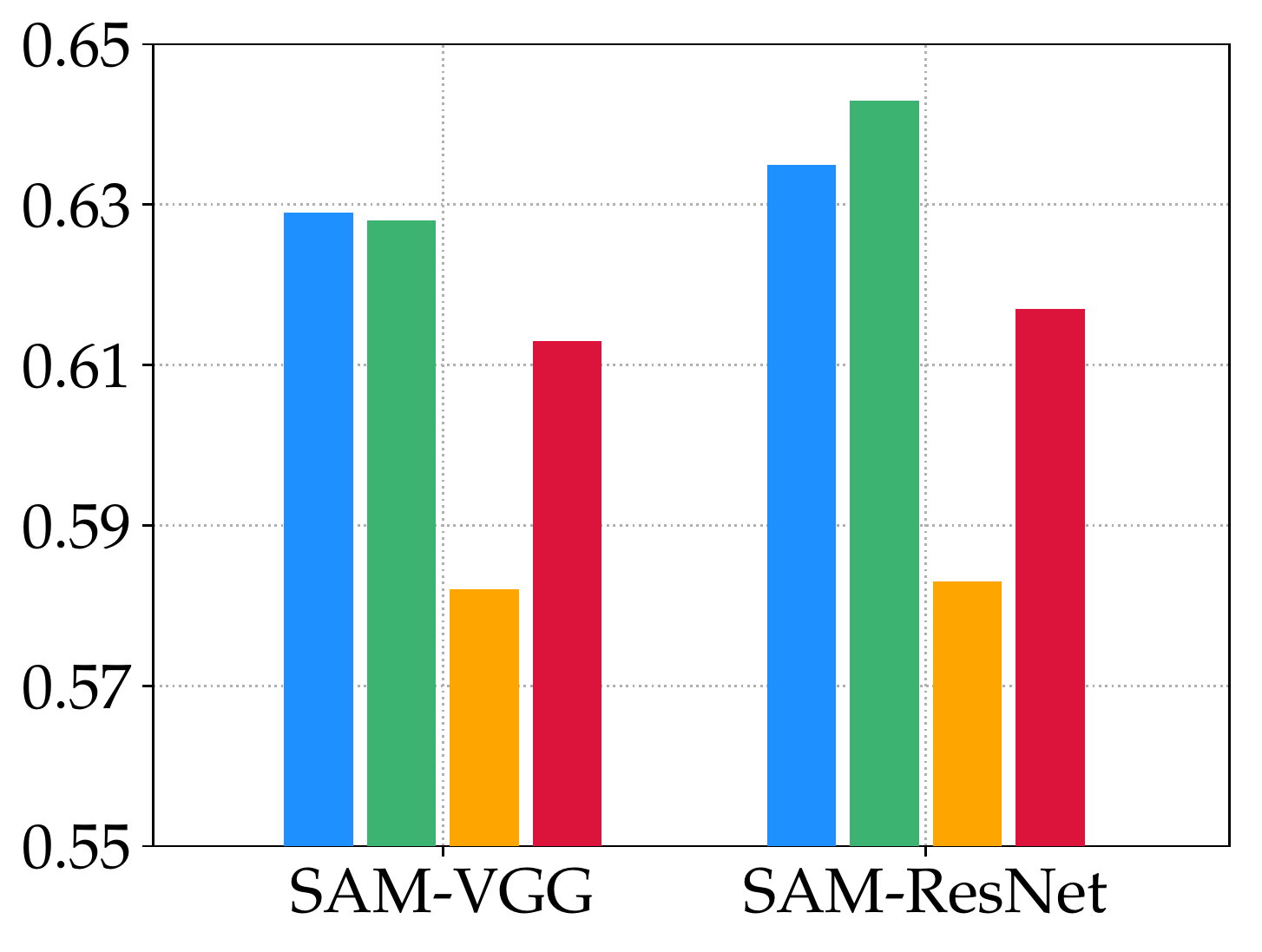}
}
\subfloat[AUC] {
	\includegraphics[width=0.475\columnwidth]{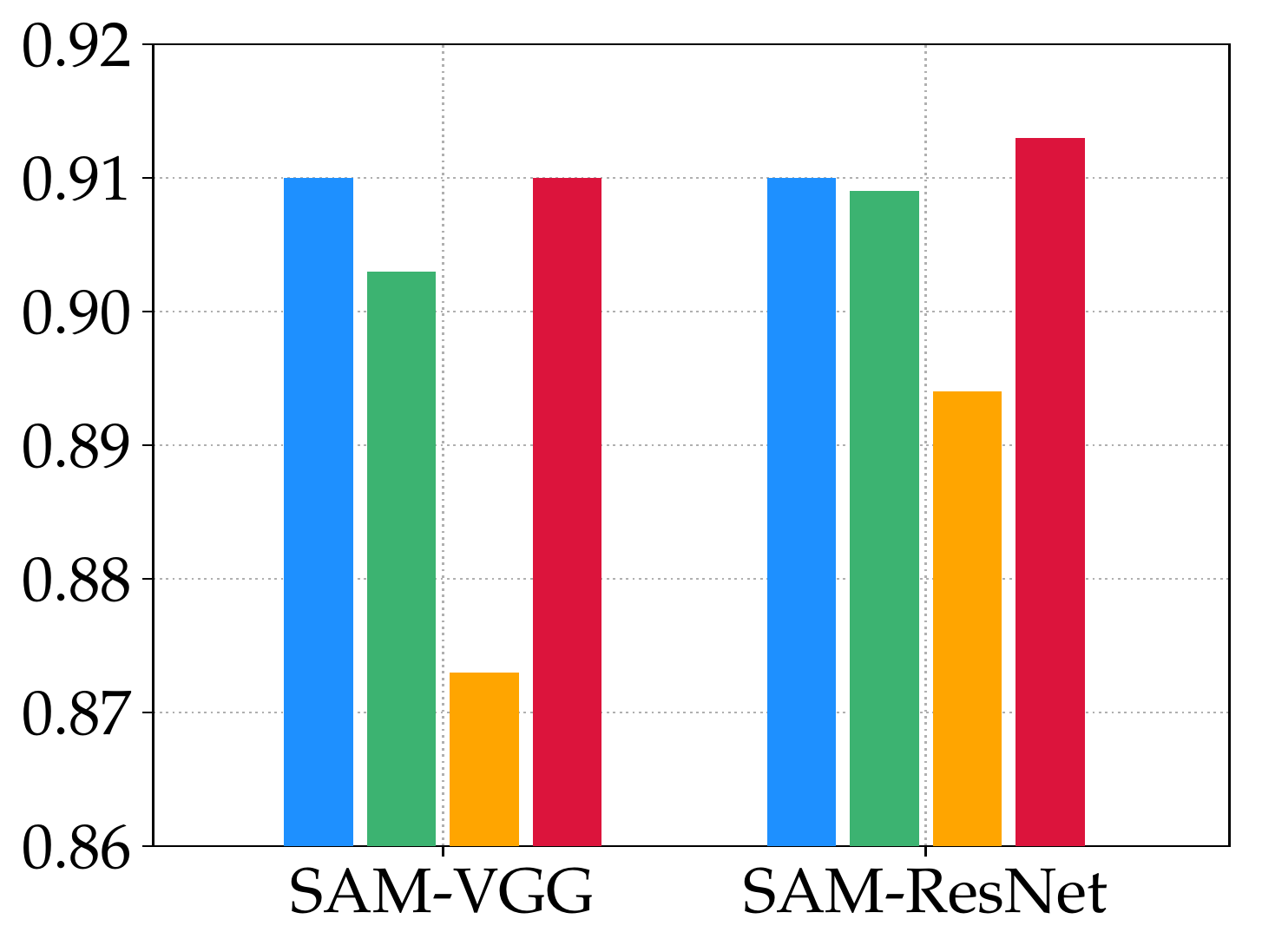}
}
\subfloat[NSS] {
	\includegraphics[width=0.475\columnwidth]{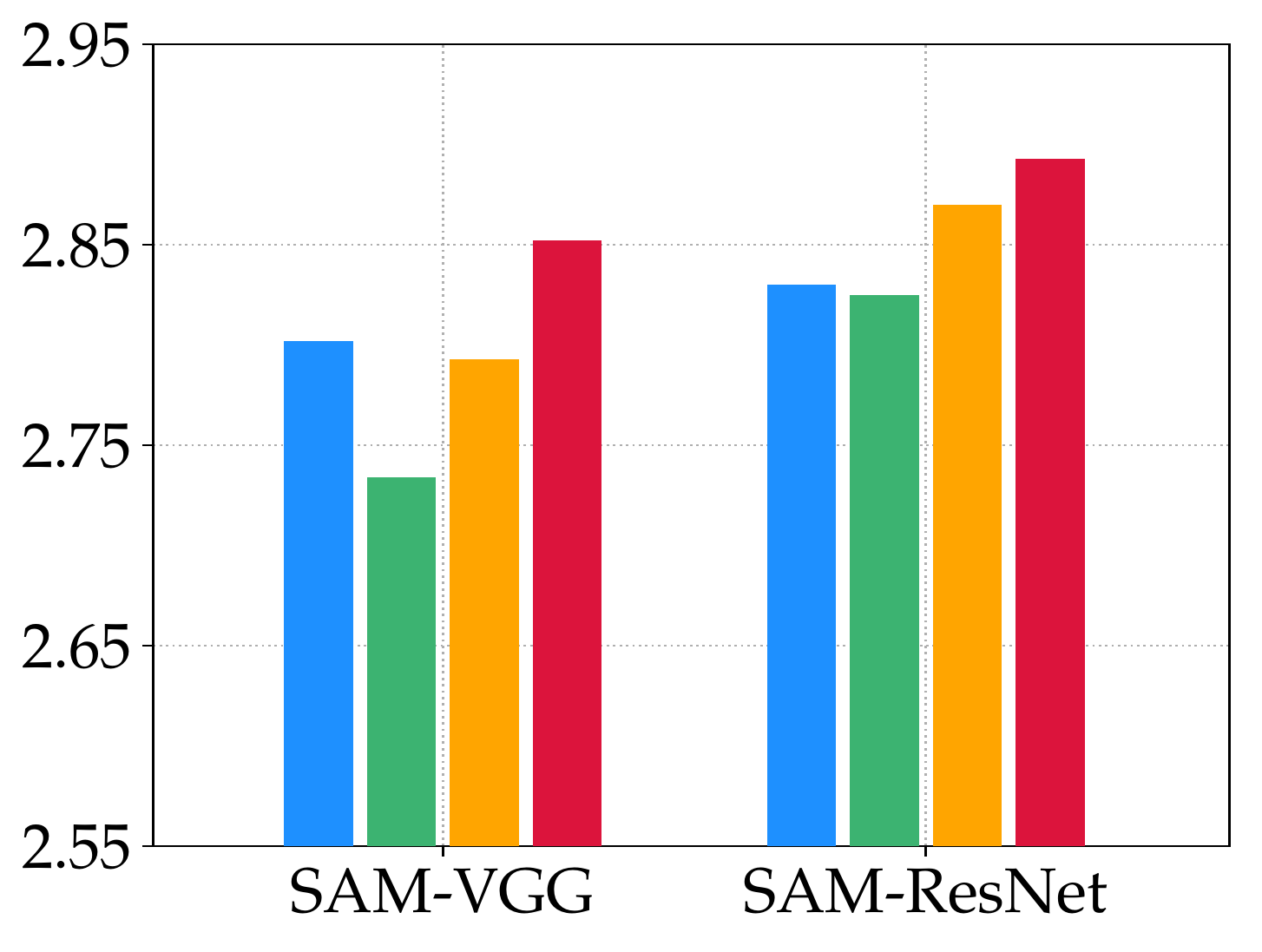}
}
\\
\small{CAT2000} \\
\subfloat[CC] {
	\includegraphics[width=0.475\columnwidth]{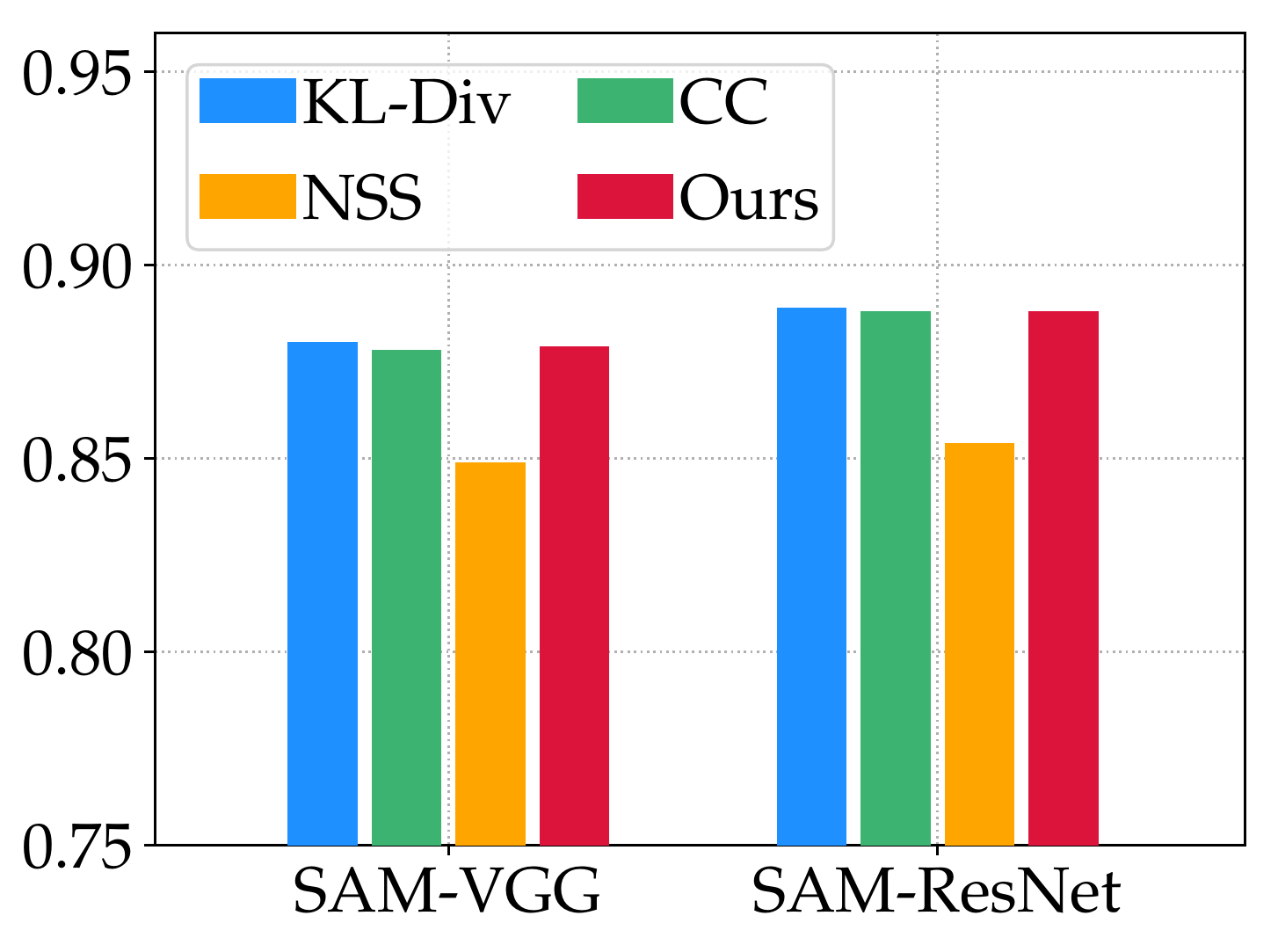}
}
\subfloat[sAUC] {
	\includegraphics[width=0.475\columnwidth]{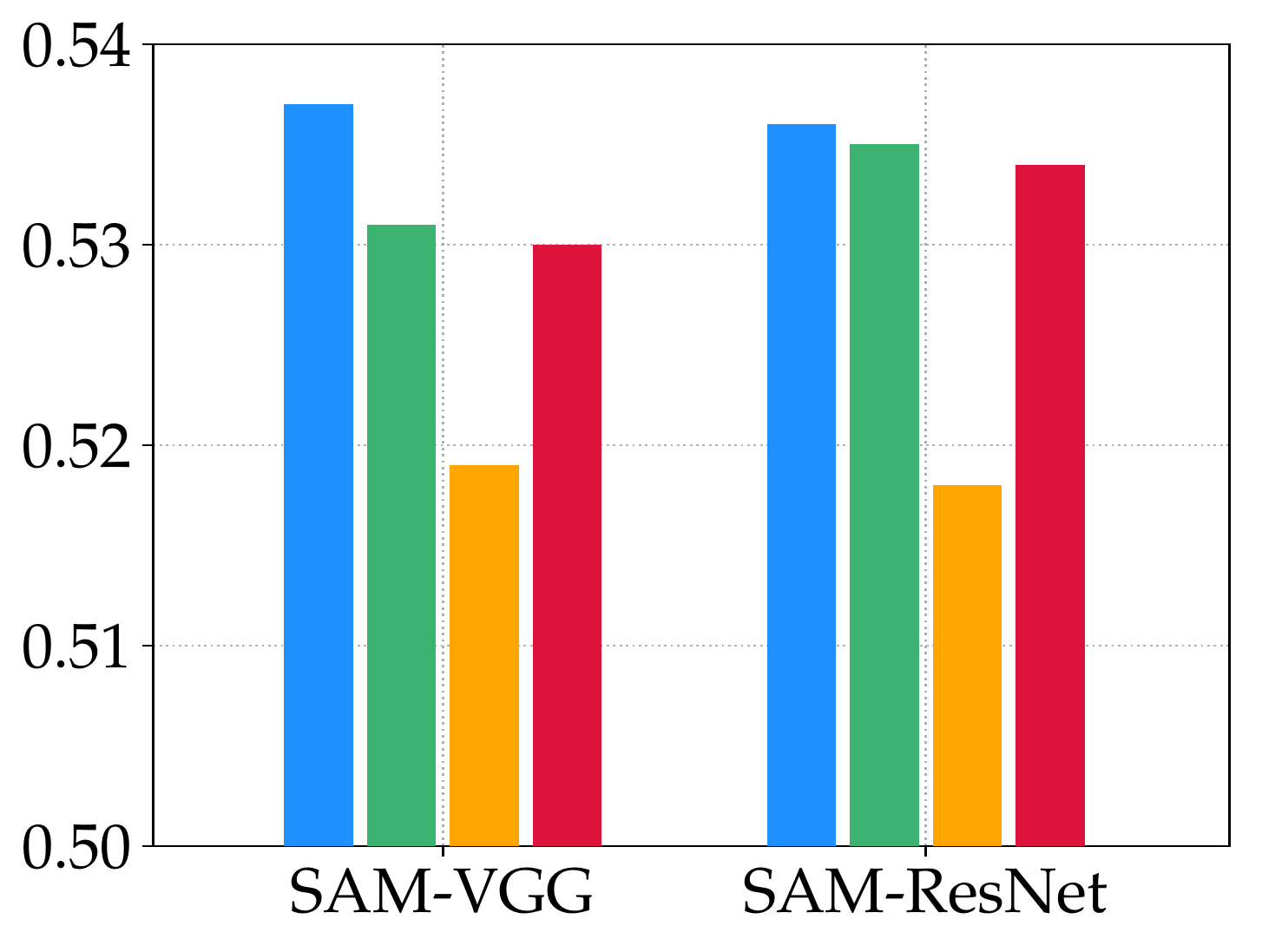}
}
\subfloat[AUC] {
	\includegraphics[width=0.475\columnwidth]{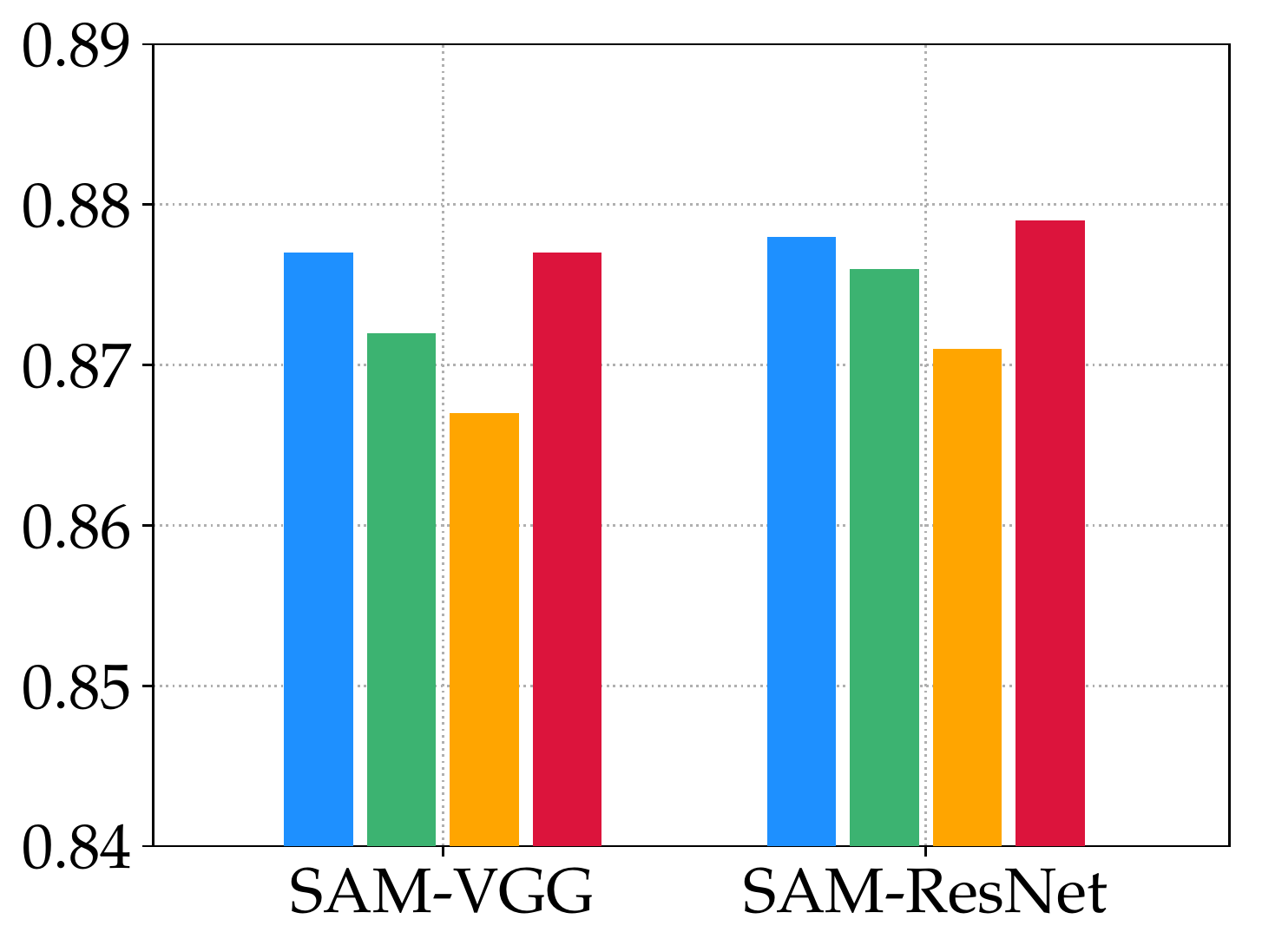}
}
\subfloat[NSS] {
	\includegraphics[width=0.475\columnwidth]{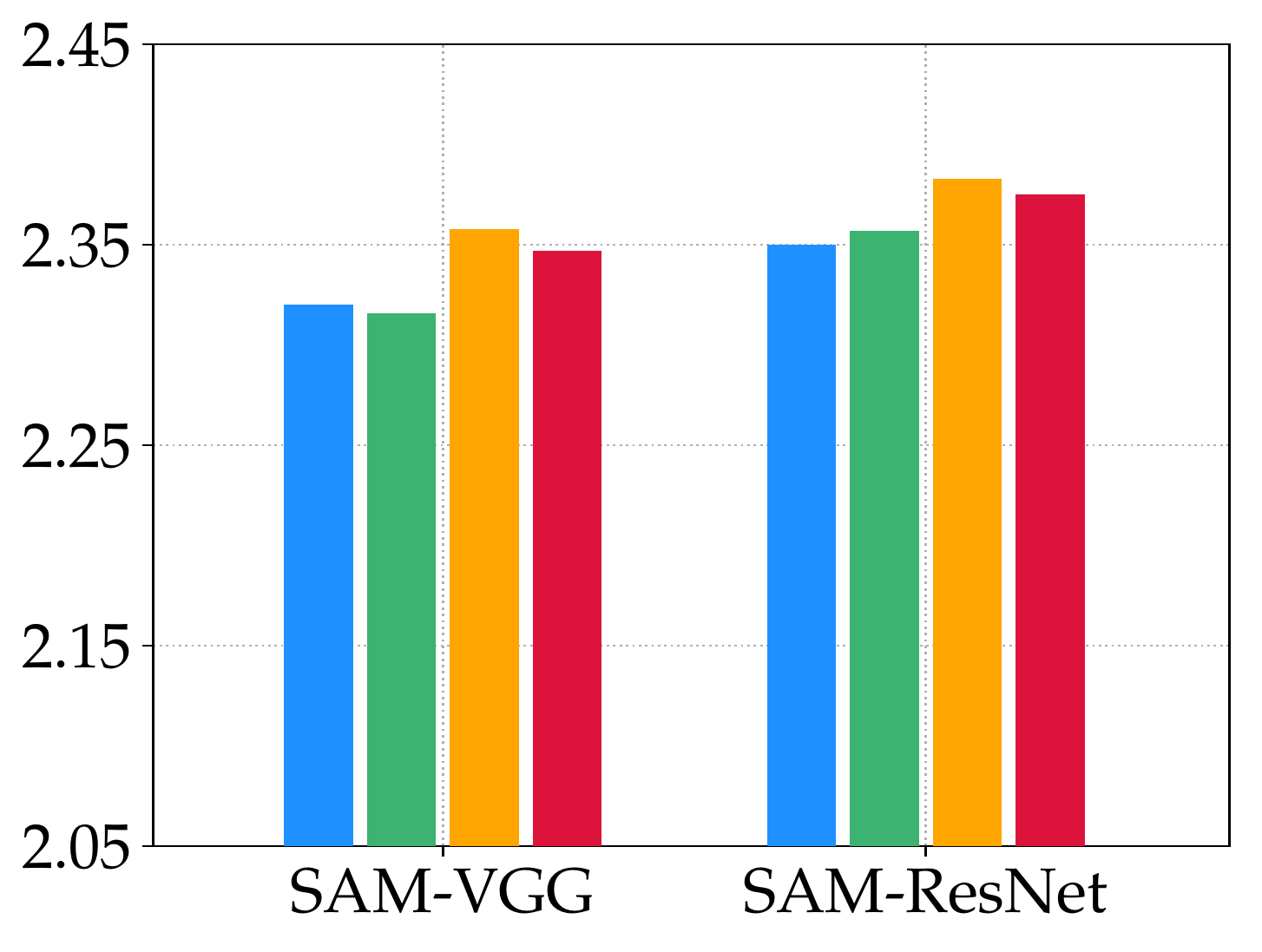}
}
\caption{Comparison between the proposed loss function and its components used individually as loss functions. We report results for both SAM-VGG and SAM-ResNet on SALICON~\cite{jiang2015salicon} (plots a-d), MIT1003~\cite{judd2009learning} (plots e-h) and CAT2000~\cite{CAT2000} (plots i-l) validation sets. Plots on the same row correspond to a different evaluation metric (CC, sAUC, AUC and NSS). The four color bars represent the loss functions used. As it can be observed, our loss function achieves the best balance between metrics.}
\label{fig:loss}
\end{figure*}

\subsection{Datasets} 
For training and testing our model, we use four of the most popular saliency datasets which differ in terms of both image content and experimental settings. \\
- SALICON~\cite{jiang2015salicon}: This is the largest available dataset for saliency prediction. It contains 10,000 training images, 5,000 validation images and 5,000 testing images, taken from the Microsoft COCO dataset~\cite{lin2014microsoft}. Eye fixations are simulated with mouse movements: as shown in~\cite{jiang2015salicon}, there is a high degree of similarity between mouse-contingent saliency annotations and fixations recorded with eye-tracking systems. Groundtruth maps of the test set are not publicly available and predictions must be submitted to the SALICON challenge website\footnote{\href{https://competitions.codalab.org/competitions/3791?secret_key=f8de41aa-090f-4fd1-967e-56fc52ad8456}{https://competitions.codalab.org/competitions/3791}} for evaluation. \\
- MIT1003~\cite{judd2009learning}: The MIT1003 dataset contains 1003 images coming from Flickr and LabelMe. Saliency maps have been created from eye-tracking data of 15 observers. \\
- MIT300~\cite{judd2012benchmark}: The MIT300 dataset is a collection of 300 natural images with saliency maps generated from eye-tracking data of 39 users. Saliency maps of this entire dataset are held out and we used the MIT Saliency benchmark~\cite{mit-saliency-benchmark} for evaluating our predictions. To test our network on this dataset, we fine-tune it on images of the MIT1003 randomly split in training and validation sets. \\
- CAT2000~\cite{CAT2000}: This dataset contains 4,000 images coming from a large variety of categories such as \textit{Cartoons}, \textit{Art}, \textit{Satellite}, \textit{Low resolution images}, \textit{Indoor}, \textit{Outdoor}, \textit{Line drawings}, ect. It is composed of 20 different categories with 200 images for each of them. Saliency maps of the testing set, composed by 2,000 images, are not available and we submitted our saliency maps to the MIT Saliency benchmark~\cite{mit-saliency-benchmark} for evaluation.

\subsection{Evaluation Metrics} 
A large variety of metrics to evaluate saliency prediction models exist and the main difference between them concerns the ground-truth representation. In fact, saliency evaluation metrics can be categorized in location-based and distribution-based metrics~\cite{riche2013saliency,salMetrics_Bylinskii,kummerer2015information}. The first category considers saliency maps at discrete fixation locations, while the second treats both ground-truth fixation maps and predicted saliency maps as continuous distributions. 

The most widely used location-based metrics are the Area under the ROC curve, in its different variants of Judd (AUC) and shuffled (sAUC), and the Normalized Scanpath Saliency (NSS). The AUC metrics do not penalize low-valued false positives giving a high score for high-valued predictions placed at fixated locations and ignoring the others. Besides, the sAUC is designed to penalize models that take into account the center bias present in eye fixations. The NSS, instead, is sensitive in an equivalent manner to both false positives and false negatives.

For the distribution-based category, the most used evaluation metrics are the Linear Correlation Coefficient (CC), the Similarity (SIM) and the Earth Mover Distance (EMD). The CC treats both false positives and false negatives symmetrically, differently from the SIM that instead measures the intersection between two distributions and for this reason it is very sensitive to missing values. The EMD is a dissimilarity metric that penalizes false positives proportionally to the spatial distance from the groundtruth. 

\subsection{Implementation Details}
We evaluate our model on SALICON, MIT300 and CAT2000 datasets. For the first dataset, we train the network on its training set and we use the $5,000$ validation images to validate the model. For the second and the third dataset, we pre-train the network on SALICON and then fine-tune on MIT1003 dataset and CAT2000 training set respectively, as suggested by the MIT Saliency Benchmark organizers. In particular, to test our model on the MIT300 dataset, we use $903$ randomly selected images of the MIT1003 to fine-tune the network and the remaining 100 as validation set. For the CAT2000 dataset, instead, we randomly choose $1,800$ images of training set for the fine-tuning and we use the remaining $200$ ($10$ for each category) as validation set.

For the SALICON, MIT1003 and MIT300 datasets, we resize input images to $240 \times 320$. Since images from MIT1003 and MIT300 have different sizes, we apply zero padding bringing images to have an aspect ratio of 4:3 and then resize them to have the selected input size. Instead, images from CAT2000 dataset have all the same input size of $1080 \times 1920$. For this reason, we resize all images of this dataset to $180 \times 320$. 

Predictions of all datasets are slightly blurred with a Gaussian filter. After a validation process, we set the standard deviation of the Gaussian kernel to $7$. 

Weights of the Dilated Convolutional Networks are initialized with those of the VGG-16 and ResNet-50 models trained on ImageNet~\cite{russakovsky2015imagenet}. For the Attentive ConvLSTM, following the initialization proposed in~\cite{bahdanau2014neural}, we initialize the recurrent weights matrices $U_i$, $U_f$, $U_o$ and $U_c$ as random orthogonal matrices. All $W$ matrices and $U_a$ are initialized by sampling each element from the Gaussian distribution of mean 0 and variance $0.05^2$. The matrix $V_a$ and all bias vectors are initialized to zero. Weights of all other convolutional layers of our model are initialized according to~\cite{glorot2010understanding}.

\begin{figure}[t!]
\centering
\subfloat[CC] {
	\includegraphics[width=0.475\columnwidth]{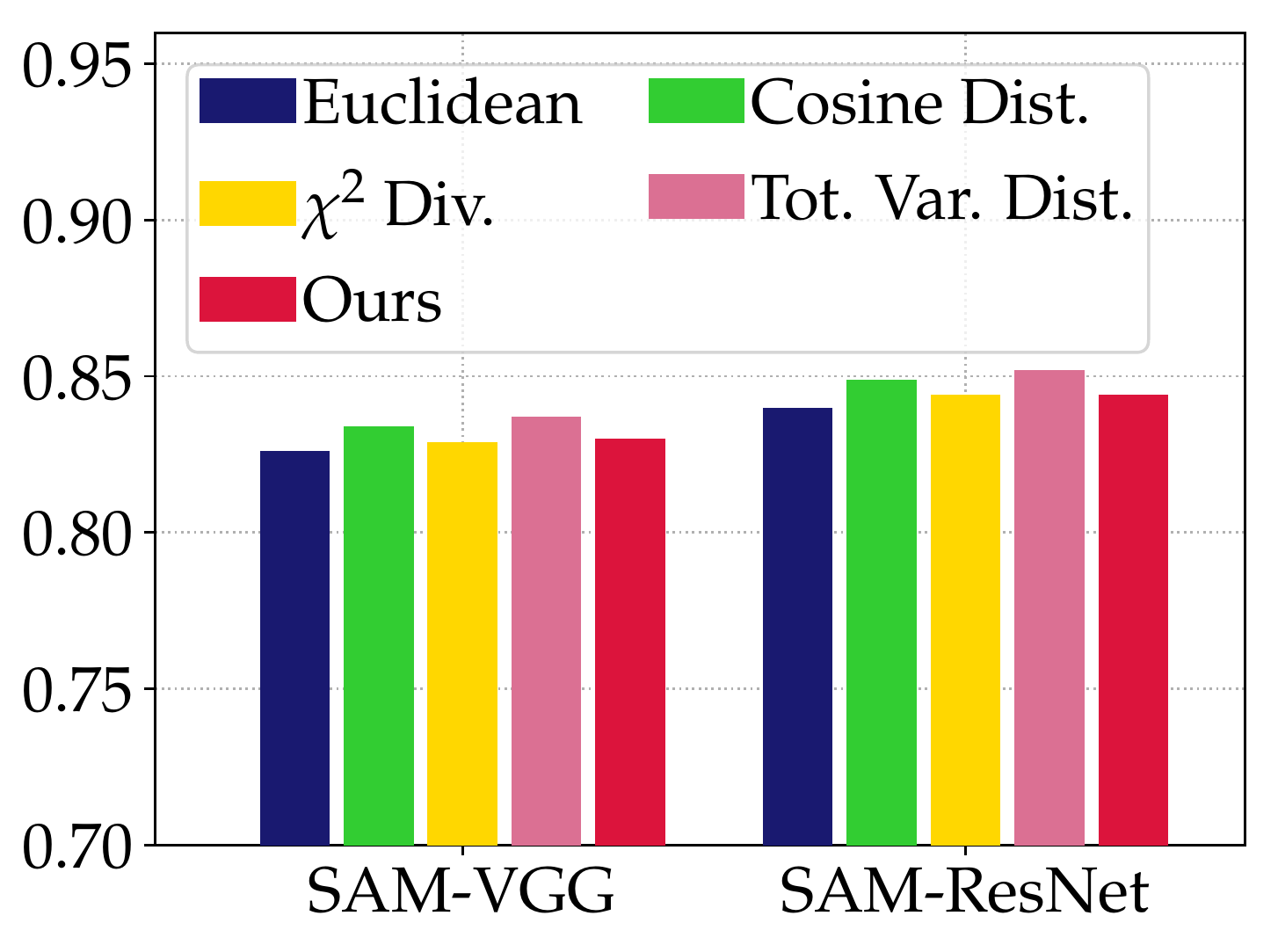}
}
\subfloat[sAUC] {
	\includegraphics[width=0.475\columnwidth]{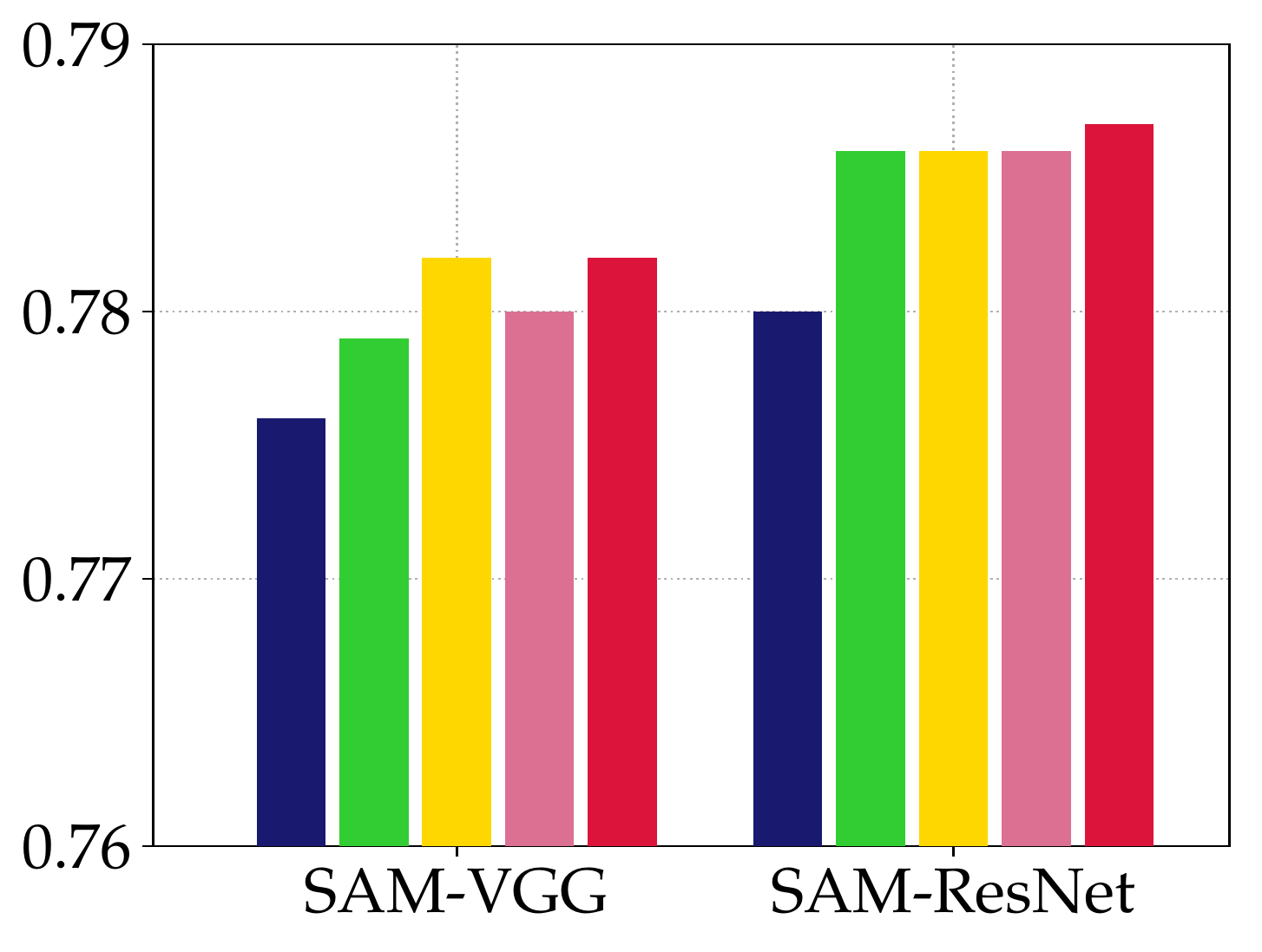}
}
\\
\subfloat[AUC] {
	\includegraphics[width=0.475\columnwidth]{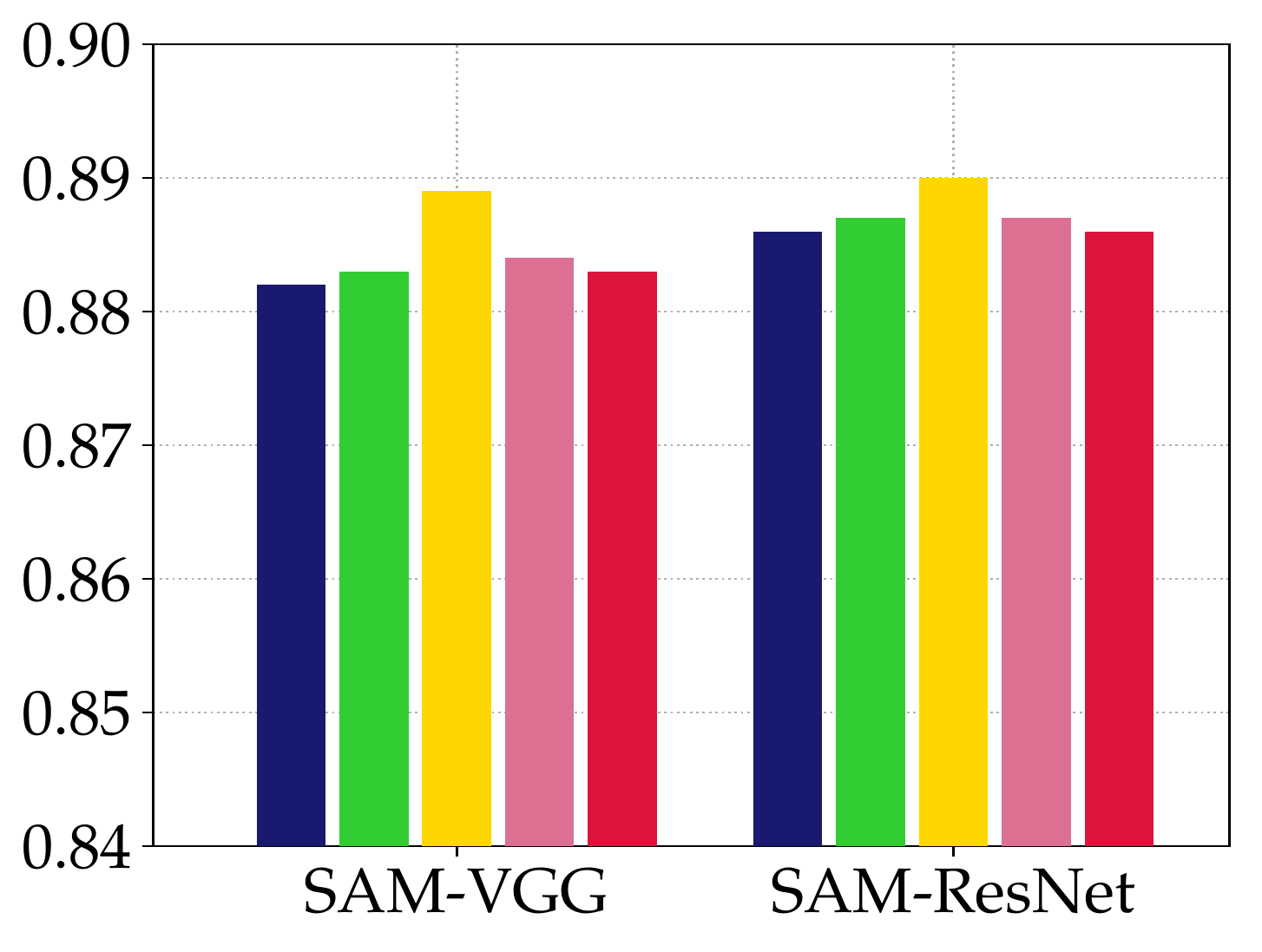}
}
\subfloat[NSS] {
	\includegraphics[width=0.475\columnwidth]{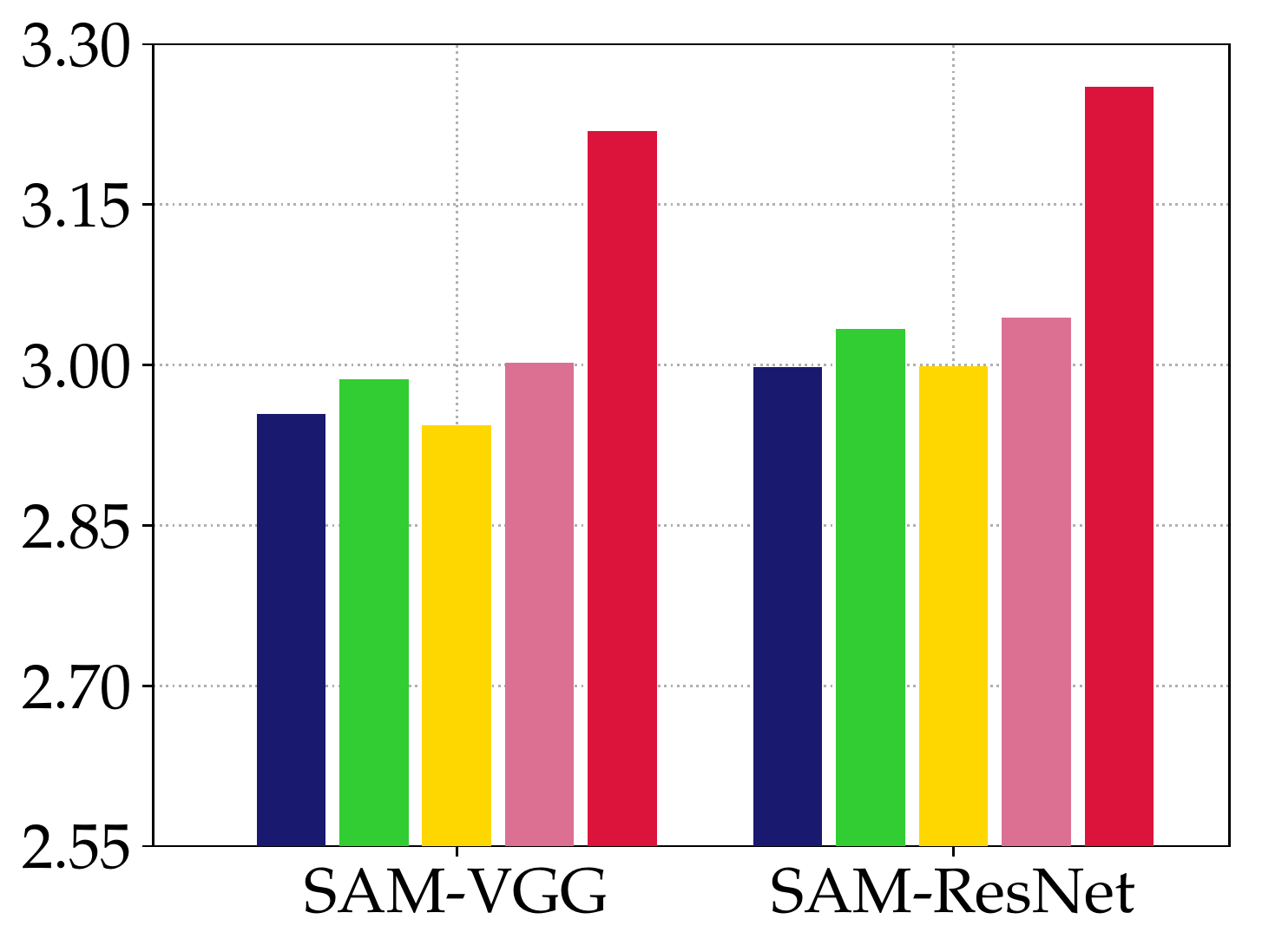}
}
\caption{Comparison between the proposed combination of saliency metrics and more traditional loss functions such as Euclidean Loss, $\chi^2$ Divergence, Cosine Distance and Total Variation Distance. Each plot corresponds to a different evaluation metric (CC, sAUC, AUC and NSS). The five color bars represent the performance of our model trained with the considered loss functions. We report results of both SAM-VGG and SAM-ResNet models on SALICON validation set~\cite{jiang2015salicon}.}
\label{fig:loss_trad}
\end{figure}

At training time, we randomly sample a minibatch containing $K$ training saliency maps, and encourage the network to minimize the proposed loss function through the RMSprop optimizer~\cite{tieleman2012lecture}. We found that a batch size of $10$ is sufficient to learn the model seamlessly. Batch normalization is preserved in the ResNet-50 part of the model, and we do not add batch normalization layers elsewhere.

Loss parameters $\alpha$, $\beta$ and $\gamma$ are respectively set to $-1$, $-2$ and $10$ balancing the contribution of each loss function. Differently from the KL-Div that is a dissimilarity metric and its value should be minimized, the CC and the NSS are to be maximized to predict better saliency maps. To this end, we set $\alpha$ and $\beta$ as negative weights. The choice of these balance weights is driven by the goal of having good results on all evaluation metrics and by taking into account the numerical range that the single metrics have at convergence.

During the training phase, we set the initial learning rate to $10^{-5}$ and we decrease it by a factor of $10$ every two epochs for the model based on the ResNet, and every three epochs for that based on the VGG network. 

\begin{table*}[t]
\renewcommand{\arraystretch}{1.3}
\centering
\caption{Ablation analysis of SAM-VGG and SAM-ResNet models on SALICON~\cite{jiang2015salicon}, MIT1003~\cite{judd2009learning} and CAT2000~\cite{CAT2000} validation sets.}
\label{tab:ablation}
\begin{small}
\begin{tabular}{|c|l|cccc|cccc|}
\hline \multirow{2}{*}{Dataset} & \multirow{2}{*}{Model} & \multicolumn{4}{c|}{\textbf{SAM-VGG}} & \multicolumn{4}{c|}{\textbf{SAM-ResNet}} \\ 
& & CC 	& sAUC 	& AUC	& NSS & CC 	& sAUC 	& AUC	& NSS\\ \hline \hline
\multirow{5}{*}{\textbf{SALICON}} & \footnotesize{Plain CNN} & 0.743 & 0.765 & 0.870 & 2.333 & 0.771 & 0.762 & 0.876 & 2.404 \\ 
 & \footnotesize{Dilated Convolutional Network} & 0.801 & \textbf{0.786} & 0.876 & 3.122 & 0.823 & 0.774 & 0.879 & 3.187 \\
 & \footnotesize{DCN + Attentive ConvLSTM} & 0.809 & 0.784 & 0.878 & 3.142 & 0.841 & 0.786 & 0.885  & 3.256 \\   		
&  \footnotesize{DCN + Learned Priors} & 0.824 & 0.782 & 0.882 & 3.209 & 0.840 & 0.784 & 0.885 & 3.235\\ 
& \footnotesize{DCN + Attentive ConvLSTM + Learned Priors} & \textbf{0.830} & 0.782 & \textbf{0.883} & \textbf{3.219} & \textbf{0.844} & \textbf{0.787} & \textbf{0.886} & \textbf{3.260} \\ \hline 
\multirow{5}{*}{\textbf{MIT1003}} &  \footnotesize{Plain CNN}  &  0.638 & \textbf{0.625} & 0.889 & 2.147 &  0.667 & \textbf{0.631} &  0.895  & 2.255 \\ 
 &  \footnotesize{Dilated Convolutional Network} & 0.718 & 0.596 & 0.906 & 2.704 & 0.748 & 0.609 & 0.902 &  2.845 \\ 
 & \footnotesize{DCN + Attentive ConvLSTM}  & 0.749 &  0.601 & 0.908 & 2.812 & 0.756 & 0.613 & 0.912 & 2.860 \\ 
 &  \footnotesize{DCN + Learned Priors} & 0.750 & 0.621 &  0.908 & 2.805 & 0.746 &  0.613 &  0.908 & 2.816 \\ 
& \footnotesize{DCN + Attentive ConvLSTM + Learned Priors}  &  \textbf{0.757} & 0.613 &  \textbf{0.910} &  \textbf{2.852} &  \textbf{0.768} &  0.617 & \textbf{0.913}  & \textbf{2.893} \\  \hline
\multirow{5}{*}{\textbf{CAT2000}} &  \footnotesize{Plain CNN}  & 0.751 & 0.546 & 0.862 & 1.886 & 0.819 &  \textbf{0.538} & 0.870  & 2.052 \\ 
 & \footnotesize{Dilated Convolutional Network} & 0.791 & \textbf{0.548} & 0.870 &  2.067 &  0.881 & 0.527 &  0.877  &  2.368 \\
& \footnotesize{DCN + Attentive ConvLSTM}  &  0.851 &  0.537 & 0.874 &  2.253 & 0.882 & 0.528 & 0.878  & 2.367 \\   		
& \footnotesize{DCN + Learned Priors} &  0.877 & 0.532 & 0.876 &  2.328 &  0.885 & 0.528 &  0.878  & \textbf{2.377} \\ 
& \footnotesize{DCN + Attentive ConvLSTM + Learned Priors}  & \textbf{0.879} & 0.530 & \textbf{0.877} & \textbf{2.347} & \textbf{0.888} & 0.534 & \textbf{0.879} & 2.375 \\ \hline
\end{tabular}
\end{small}
\end{table*}

\section{Experimental Evaluation}
In this section we perform analyses and experiments to validate the contribution of each component of the network. We also show quantitative and qualitative comparisons with other state of the art models.

\begin{figure}[t]
\centering
\setlength\tabcolsep{1pt}
\begin{tabular}{cccc}
 \small{(a)} & \small{(b)} & \small{(c)} & \small{Groundtruth}  \\
\includegraphics[width=0.22\columnwidth]{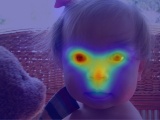}&
\includegraphics[width=0.22\columnwidth]{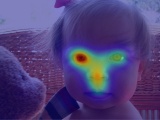}&
\includegraphics[width=0.22\columnwidth]{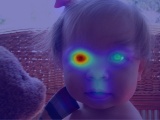}&
\includegraphics[width=0.22\columnwidth]{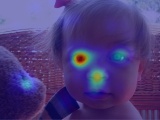}\\
\includegraphics[width=0.22\columnwidth]{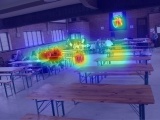}&
\includegraphics[width=0.22\columnwidth]{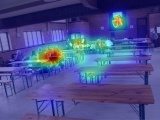}&
\includegraphics[width=0.22\columnwidth]{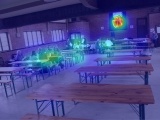}&
\includegraphics[width=0.22\columnwidth]{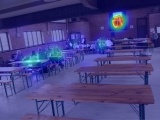}\\
\includegraphics[width=0.22\columnwidth]{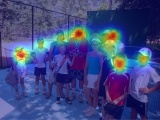}&
\includegraphics[width=0.22\columnwidth]{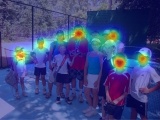}&
\includegraphics[width=0.22\columnwidth]{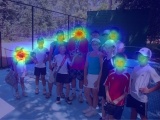}&
\includegraphics[width=0.22\columnwidth]{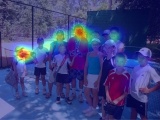}\\
\includegraphics[width=0.22\columnwidth]{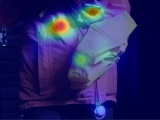}&
\includegraphics[width=0.22\columnwidth]{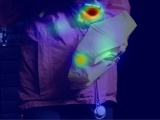}&
\includegraphics[width=0.22\columnwidth]{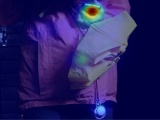}&
\includegraphics[width=0.22\columnwidth]{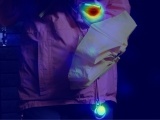}\\
\end{tabular}
\caption{Examples of saliency maps predicted by the DCN (a), the DCN with the Attentive ConvLSTM (b), and the DCN with the Attentive ConvLSTM and learned priors (c) compared with the groundtruth (d). Images are from SALICON validation set~\cite{jiang2015salicon}.}
\label{fig:images}
\end{figure}

\subsection{Comparison between different loss functions}\label{sec:loss}
In Fig.~\ref{fig:loss} we compare results obtained by using single loss functions (KL-Div, CC, NSS) and our combination proposed in Section~\ref{sec:training}. Results are reported for both versions of our model on SALICON, MIT1003 and CAT2000 validations sets. We call SAM-VGG the model based on the VGG network and SAM-ResNet that based on the ResNet network.

As it can be seen, our combined loss achieves on average better results on all metrics. For example on the SALICON dataset, when the model is trained using the KL-Div or the CC metrics as loss function, the performances are good especially on the CC, while the model fails on the NSS. When the model is trained using the NSS metric, instead, it achieves better results only on the NSS and fails on all other metrics. A similar behaviour is also present on the MIT1003 and CAT2000 datasets where the gain in performance obtained by our loss function is particularly evident on the CC, AUC and NSS metrics, even reaching in some cases the best results.

To further validate the effectiveness of the proposed loss function, we compare it with traditional loss functions and probability distances used by other previous saliency models~\cite{kruthiventi2016saliency,cPan,jetley2016end}. Fig.~\ref{fig:loss_trad} shows the comparison between our combination of saliency metrics and four other loss functions: the Euclidean loss, the Cosine Distance, the $\chi^2$ Divergence and the Total Variation Distance. Also in this case, our loss function achieves a better balance among all metrics. The gap with respect to all other traditional losses is particularly evident on the NSS metric, while, on all other metrics, the proposed combined loss, if it does not reach the best results, it is very close to them.

Overall, our combined loss reaches competitive results on all metrics differently from the other loss functions. For this reason, results of all following experiments are obtained by training the network with our combination of loss.

\subsection{Model Ablation Analysis}
We evaluate the contribution of each component of the architecture, on SALICON, MIT1003 and CAT2000 validation sets. To this end, we construct five different variations: the plain CNN architecture without the last fully convolutional layer (as a baseline), the Dilated Convolutional Network (DCN), the DCN with the proposed ConvLSTM model, the DCN with the proposed learned priors module and the final version of our model with all its components.

Table~\ref{tab:ablation} shows the results of the ablation analysis using both versions of our model on three different datasets. The results emphasize that the overall architecture is able to predict better saliency maps in both SAM-VGG and SAM-ResNet variants and each proposed component gives an important contribution to the final performance on all considered datasets. In particular, on the SALICON dataset, it can be seen that there is a constant improvement on all metrics. For example, the VGG baseline achieves a result of $0.743$ in terms of CC, while the DCN achieves a relative improvement of $\frac{0.801-0.743}{0.743}=7.8\%$. This result is further improved by $1\%$ when adding the Attentive ConvLSTM or by $2.9\%$ when adding the learned priors. The overall architecture adds an important improvement of $2.6\%$ to the DCN with the Attentive ConvLSTM and $0.7\%$ to the DCN with learned priors. 
The ResNet baseline, instead, achieves a CC result of $0.771$ that is improved by a $6.7\%$ when adding the dilated convolutions. The Attentive ConvLSTM and the learned priors respectively add an improvement of $2.2\%$ and $2.1\%$. These results are further improved using the overall architecture with all proposed components by $0.4\%$ and $0.5\%$.

It is also noteworthy that, with our pipeline, a VGG-based network and a ResNet-based network achieve almost the same performance, so one of the two models can be equally chosen according to speed and memory allocation needs, without considerably affecting prediction performance.

Figure~\ref{fig:images} shows some qualitative examples of saliency maps predicted by our SAM-ResNet model and by only some of its main components with respect to the groundtruth. As it can be seen, there is a constant improvement of predictions which, by adding our key components, are more qualitatively similar to the groundtruth.

\begin{table*}[t]
\renewcommand{\arraystretch}{1.3}
\centering
\caption{Comparison results between our learned priors and that proposed in~\cite{mlnet2016} on SALICON~\cite{jiang2015salicon}, MIT1003~\cite{judd2009learning} and CAT2000~\cite{CAT2000} validation sets.}
\label{tab:priors}
\begin{small}
\begin{tabular}{|l|cccc|cccc|cccc|}
\hline
& \multicolumn{4}{c|}{\textbf{SALICON}} & \multicolumn{4}{c|}{\textbf{MIT1003}} & \multicolumn{4}{c|}{\textbf{CAT2000}} \\ 
& CC & sAUC & AUC & NSS & CC & sAUC & AUC & NSS & CC & sAUC & AUC & NSS \\ \hline \hline        
SAM-VGG (prior of~\cite{mlnet2016}) 
& 0.811 & \textbf{0.783} & 0.878 & 3.150 
& 0.738 & 0.610 & 0.908 & 2.754
& 0.845 & \textbf{0.539} & 0.874 & 2.233 \\ 
SAM-VGG (learned priors) 
& \textbf{0.830} & 0.782 & \textbf{0.883} & \textbf{3.219} 
& \textbf{0.757} & \textbf{0.613} & \textbf{0.910} & \textbf{2.852}
& \textbf{0.879} & 0.530 & \textbf{0.877} & \textbf{2.347} \\ \hline 
SAM-ResNet (prior of~\cite{mlnet2016}) 	
& 0.840 & 0.785 & 0.884 & 3.249 
& 0.766 & 0.609 & 0.912 & \textbf{2.899}
& 0.886 & 0.528 & 0.878 & \textbf{2.386} \\ 
SAM-ResNet (learned priors) 
& \textbf{0.844} & \textbf{0.787} & \textbf{0.886} & \textbf{3.260} 
& \textbf{0.768} & \textbf{0.617} & \textbf{0.913} & 2.893
& \textbf{0.888} & \textbf{0.534} & \textbf{0.879} & 2.375 \\ \hline 
\end{tabular}
\end{small}
\end{table*}

\begin{table}[t]
\renewcommand{\arraystretch}{1.3}
\centering
\caption{Results on SALICON validation set~\cite{jiang2015salicon} when using the output of the Attentive ConvLSTM module at different timesteps as input of the rest of the model.}
\label{tab:tick}
\begin{small}
\begin{tabular}{|c|c|cccc|}
\hline
		& T & CC 		& sAUC 	& AUC	& NSS	\\ \hline \hline        
\multirow{4}{*}{SAM-VGG} & 1 & 0.821 & 0.777 & \textbf{0.884} & 3.168 \\ 
 & 2 & 0.827 & 0.777 & 0.883 & 3.224 \\ 
 & 3 & 0.828 & 0.781 & 0.883 & \textbf{3.226} \\ 
 & 4 & \textbf{0.830} & \textbf{0.782} & 0.883 & 3.219 \\ 
\hline
\multirow{4}{*}{SAM-ResNet} & 1 & 0.785 & 0.737 & 0.879 & 3.050 \\ 
 & 2 & 0.829 & 0.764 & \textbf{0.886} & 3.214 \\ 
 & 3 & 0.842 & 0.779 & \textbf{0.886} & 3.256 \\
 & 4 & \textbf{0.844} & \textbf{0.787} & \textbf{0.886} & \textbf{3.260}  \\ \hline
\end{tabular}
\end{small}
\end{table}

\subsection{Contribution of the attentive model and learned priors}\label{sec:exp_priors}
Table~\ref{tab:tick} reports the performance of our model when using the output of the Attentive ConvLSTM module at different timesteps as input for the rest of the model. Results clearly show that the refinement carried out by the Attentive model results in better performance. No further significant improvements were observed for $t > 4$: while CC, sAUC and AUC almost saturated, NSS slightly decreased after four iterations.

To assess the effectiveness of our prior learning strategy, we compare it with the approach in~\cite{mlnet2016}, in which a low resolution prior map is learned and applied element-wise to the predicted saliency map, after performing bilinear upsampling. We chose to compare our solution to that in~\cite{mlnet2016} because it is the only other attempt to incorporate the center bias in a deep learning model without the use of hand-crafted prior maps.
Results on SALICON, MIT1003 and CAT2000 validation sets are reported in Table~\ref{tab:priors}. Using multiple Gaussian learned priors, instead of learning an entire prior map, with no pre-defined structure, shows to be beneficial according to all metrics.

\subsection{Comparison with state of the art}
We quantitatively compare our method with state of the art models on SALICON, MIT300 and CAT2000 test sets. Not all saliency methods report experimental results on all considered datasets. For this reason, comparison methods are different depending on each dataset. We decide to sort model performances by the NSS metric as suggested by the MIT Saliency Benchmark~\cite{mit-saliency-benchmark,salMetrics_Bylinskii,kummerer2015information}.

Table~\ref{tab:salicon} shows the results on the SALICON dataset in terms of CC, sAUC, AUC and NSS. As it can be observed, our SAM-ResNet solution outperforms all competitors by a big margin especially on CC and NSS metrics and obtains the best result also on the sAUC. In particular, our method overcomes the other ResNet-based model~\cite{liu2016deep} with an improvement of $1.5\%$ according to NSS metric, $1.3\%$ and $0.4\%$ according to CC and sAUC. For a fair comparison with other methods, we also include the results achieved by our SAM-VGG model. The improvement with respect all other VGG-based methods is even more significant than that obtained by the SAM-ResNet model. In detail, our SAM-VGG overcomes all other VGG-based methods with an improvement of $12.7\%$ and $5.6\%$ according to NSS and CC metrics.

\begin{table}[tb]
\renewcommand{\arraystretch}{1.3}
\centering
\caption{Comparison results on SALICON test set~\cite{jiang2015salicon}. The results in bold indicate the best performing method on each evaluation metric. (*) indicates citations to non-peer reviewed texts. Methods are sorted by the NSS metric.}
\label{tab:salicon}
\begin{small}
\begin{tabular}{|l|cccc|}
\hline
		& CC 		& sAUC 	& AUC	& NSS	\\ \hline \hline       
\textbf{SAM-ResNet} 	& \textbf{0.842} & \textbf{0.779} & 0.883 & \textbf{3.204} \\ \hline
DSCLRCN~\cite{liu2016deep} & 0.831	& 0.776	& 0.884 & 3.157 \\ \hline
\textbf{SAM-VGG} 	& 0.825 & 0.774 & 0.881 & 3.143 \\ \hline
ML-Net~\cite{mlnet2016}		& 0.743 & 0.768	& 0.866 & 2.789	\\ \hline
MixNet~\cite{dodge2017visual} & 0.730	& 0.771	 & 0.861	& 2.767 \\ \hline
SU~\cite{kruthiventi2016saliency} & 0.780 	& 0.760 & 0.880 & 2.610 \\ \hline
SalGAN~\cite{pan2017salgan} (*) & 0.781 & 0.772 & 0.781 & 2.459 \\ \hline 
SalNet~\cite{cPan}	& 0.622	& 0.724	 & 0.858	& 1.859 \\ \hline
DeepGazeII~\cite{kummerer2016deepgaze} & 0.509 & 0.761 & \textbf{0.885} & 1.336 \\ \hline
\end{tabular}
\end{small}
\end{table}

With the proposed model, we have also participated to the LSUN Challenge 2017, where we reached the first place on the saliency prediction task\footnote{\url{https://competitions.codalab.org/competitions/17136}}.

The results on MIT300 and CAT2000 datasets are respectively reported in Tables~\ref{tab:mit300} and~\ref{tab:cat2000}. Our method achieves state of the art results on all metrics, except for the sAUC, on the CAT2000 dataset surpassing other methods by an important margin especially on SIM, CC, NSS and EMD metrics. On the MIT300 dataset, instead, we obtain results very close to the best ones. 

Our model does not obtain a big gain in performance on AUC metrics. This can be explained considering that the AUC metrics are primarily based on true positives without significantly penalizing false positives. For this reason, hazy or blurred saliency maps like the ones predicted by~\cite{kummerer2016deepgaze} achieve high AUC values~\cite{borji2013analysis,zhao2011learning}, despite being visually very different from the groundtruth annotations, as we will show in the following.

\begin{figure*}[t]
\centering
\setlength\tabcolsep{0.5pt}
\begin{footnotesize}
\begin{tabular}{m{1.25cm}m{1.25cm}m{1.25cm}m{1.25cm}m{1.25cm}m{1.25cm}m{1.25cm}m{0.07cm}m{1.25cm}m{1.25cm}m{1.25cm}m{1.25cm}m{1.25cm}m{1.25cm}m{1.25cm}}
\centering Image & \centering \cite{kummerer2016deepgaze} & \centering \cite{cPan} & \centering \cite{mlnet2016} & \centering \textbf{SAM-VGG} & \centering \textbf{SAM-ResNet} & \centering GT & & \centering Image & \centering \cite{vig2014large} & \centering \cite{liu2015predicting} & \centering \cite{mlnet2016} & \centering \textbf{SAM-VGG} & \centering \textbf{SAM-ResNet} & \centering GT \arraybackslash \\
\includegraphics[width=0.067\textwidth]{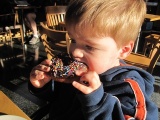}&
\includegraphics[width=0.067\textwidth]{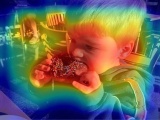}&
\includegraphics[width=0.067\textwidth]{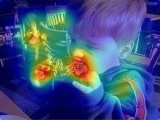}&
\includegraphics[width=0.067\textwidth]{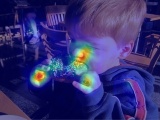}&
\includegraphics[width=0.067\textwidth]{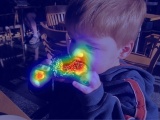}&
\includegraphics[width=0.067\textwidth]{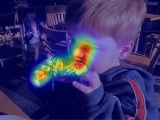}&
\includegraphics[width=0.067\textwidth]{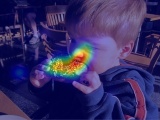}& &
\includegraphics[width=0.067\textwidth]{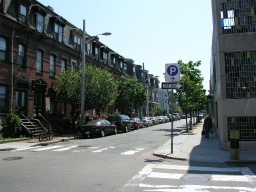}&
\includegraphics[width=0.067\textwidth]{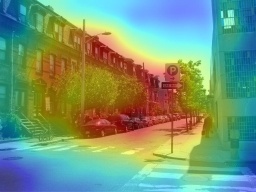}&
\includegraphics[width=0.067\textwidth]{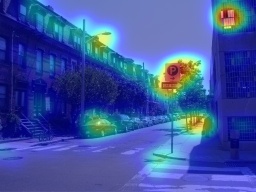}&
\includegraphics[width=0.067\textwidth]{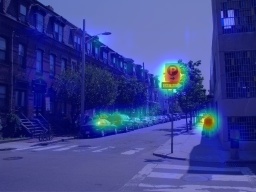}&
\includegraphics[width=0.067\textwidth]{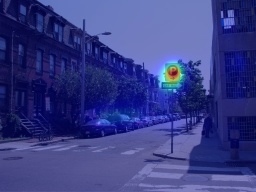}&
\includegraphics[width=0.067\textwidth]{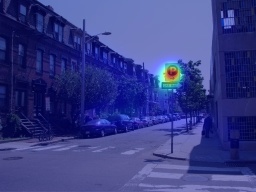}&
\includegraphics[width=0.067\textwidth]{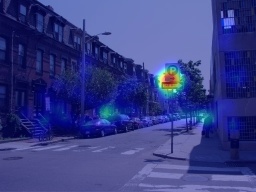}\\
\includegraphics[width=0.067\textwidth]{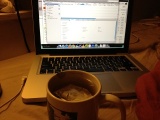}&
\includegraphics[width=0.067\textwidth]{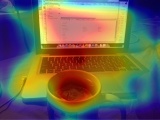}&
\includegraphics[width=0.067\textwidth]{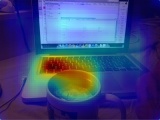}&
\includegraphics[width=0.067\textwidth]{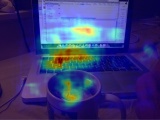}&
\includegraphics[width=0.067\textwidth]{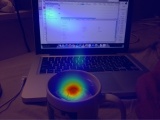}&
\includegraphics[width=0.067\textwidth]{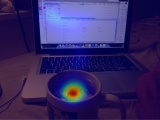}&
\includegraphics[width=0.067\textwidth]{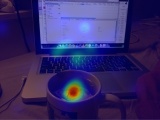}& &
\includegraphics[width=0.067\textwidth]{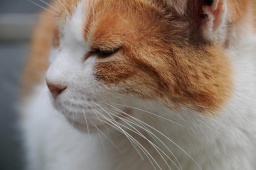}&
\includegraphics[width=0.067\textwidth]{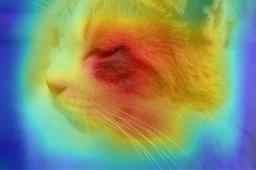}&
\includegraphics[width=0.067\textwidth]{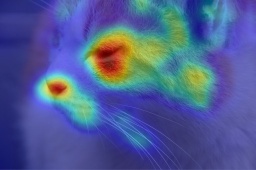}&
\includegraphics[width=0.067\textwidth]{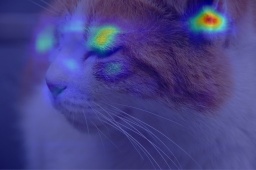}&
\includegraphics[width=0.067\textwidth]{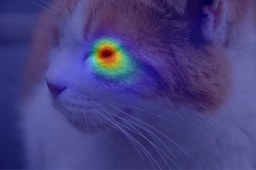}&
\includegraphics[width=0.067\textwidth]{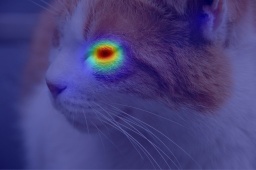}&
\includegraphics[width=0.067\textwidth]{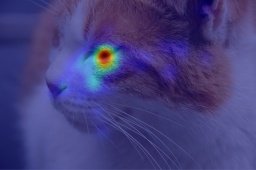}\\
\includegraphics[width=0.067\textwidth]{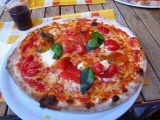}&
\includegraphics[width=0.067\textwidth]{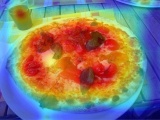}&
\includegraphics[width=0.067\textwidth]{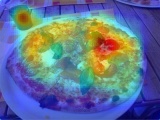}&
\includegraphics[width=0.067\textwidth]{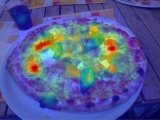}&
\includegraphics[width=0.067\textwidth]{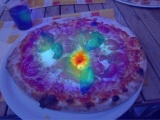}&
\includegraphics[width=0.067\textwidth]{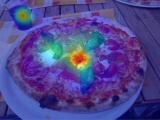}&
\includegraphics[width=0.067\textwidth]{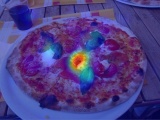}& &
\includegraphics[width=0.067\textwidth]{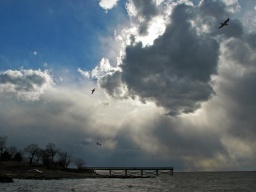}&
\includegraphics[width=0.067\textwidth]{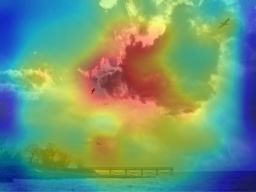}&
\includegraphics[width=0.067\textwidth]{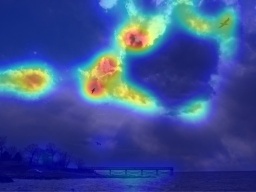}&
\includegraphics[width=0.067\textwidth]{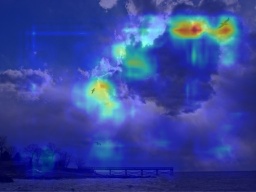}&
\includegraphics[width=0.067\textwidth]{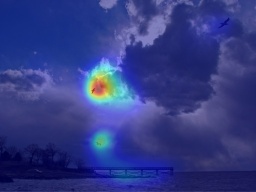}&
\includegraphics[width=0.067\textwidth]{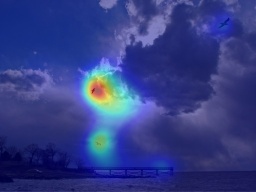}&
\includegraphics[width=0.067\textwidth]{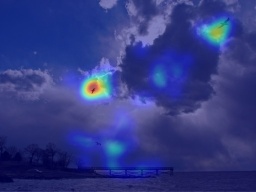}\\
\includegraphics[width=0.067\textwidth]{images/COCO_val2014_000000498747.jpg}&
\includegraphics[width=0.067\textwidth]{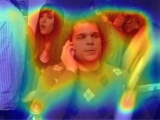}&
\includegraphics[width=0.067\textwidth]{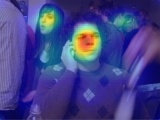}&
\includegraphics[width=0.067\textwidth]{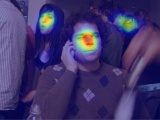}&
\includegraphics[width=0.067\textwidth]{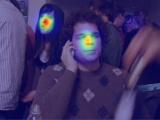}&
\includegraphics[width=0.067\textwidth]{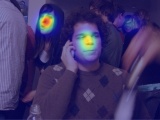}&
\includegraphics[width=0.067\textwidth]{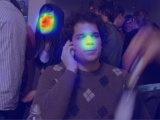}& &
\includegraphics[width=0.067\textwidth]{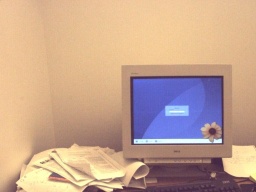}&
\includegraphics[width=0.067\textwidth]{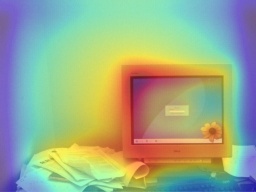}&
\includegraphics[width=0.067\textwidth]{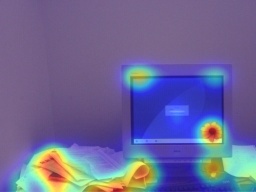}&
\includegraphics[width=0.067\textwidth]{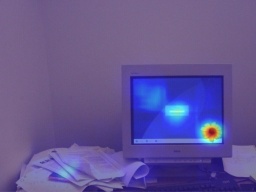}&
\includegraphics[width=0.067\textwidth]{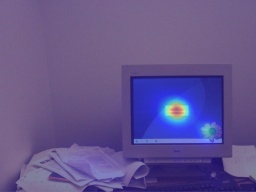}&
\includegraphics[width=0.067\textwidth]{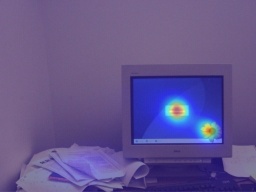}&
\includegraphics[width=0.067\textwidth]{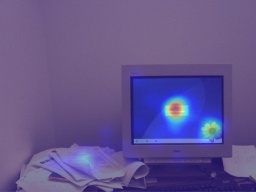}\\
\includegraphics[width=0.067\textwidth]{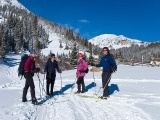}&
\includegraphics[width=0.067\textwidth]{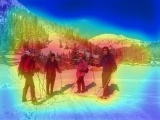}&
\includegraphics[width=0.067\textwidth]{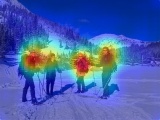}&
\includegraphics[width=0.067\textwidth]{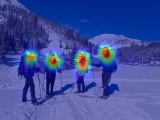}&
\includegraphics[width=0.067\textwidth]{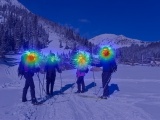}&
\includegraphics[width=0.067\textwidth]{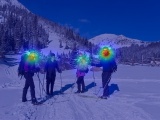}&
\includegraphics[width=0.067\textwidth]{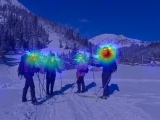}& &
\includegraphics[width=0.067\textwidth]{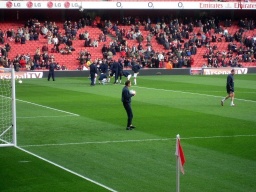}&
\includegraphics[width=0.067\textwidth]{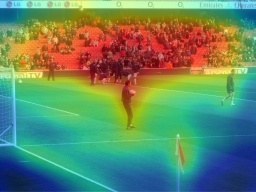}&
\includegraphics[width=0.067\textwidth]{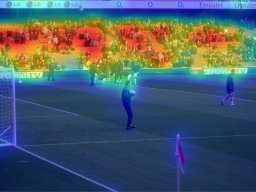}&
\includegraphics[width=0.067\textwidth]{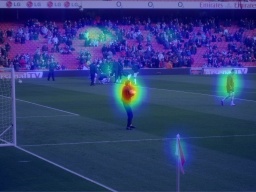}&
\includegraphics[width=0.067\textwidth]{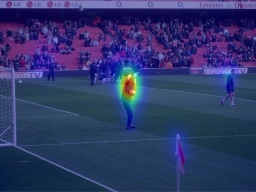}&
\includegraphics[width=0.067\textwidth]{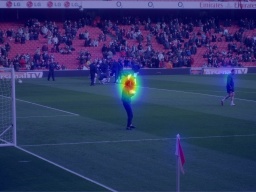}&
\includegraphics[width=0.067\textwidth]{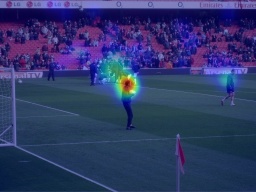}\\
\includegraphics[width=0.067\textwidth]{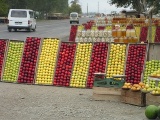}&
\includegraphics[width=0.067\textwidth]{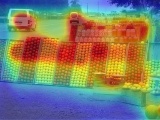}&
\includegraphics[width=0.067\textwidth]{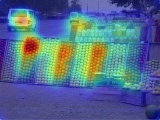}&
\includegraphics[width=0.067\textwidth]{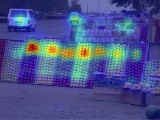}&
\includegraphics[width=0.067\textwidth]{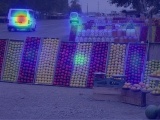}&
\includegraphics[width=0.067\textwidth]{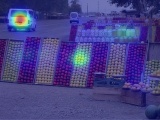}&
\includegraphics[width=0.067\textwidth]{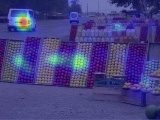}& &
\includegraphics[width=0.067\textwidth]{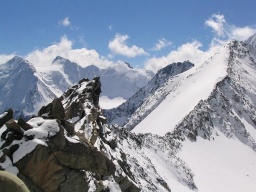}&
\includegraphics[width=0.067\textwidth]{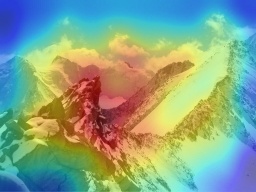}&
\includegraphics[width=0.067\textwidth]{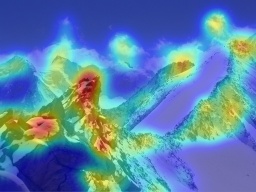}&
\includegraphics[width=0.067\textwidth]{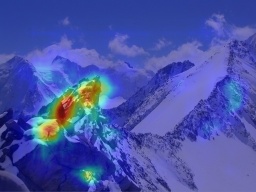}&
\includegraphics[width=0.067\textwidth]{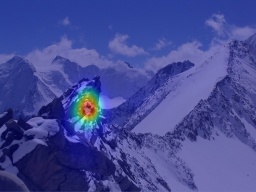}&
\includegraphics[width=0.067\textwidth]{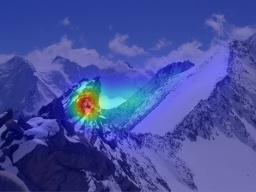}&
\includegraphics[width=0.067\textwidth]{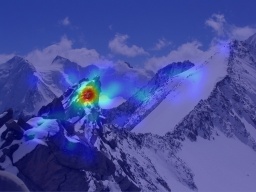}\\
\end{tabular}
\end{footnotesize}
\caption{Qualitative results and comparison with other state of the art models. Left images are from SALICON validation set~\cite{jiang2015salicon}, while right images are from MIT1003 validation set~\cite{judd2009learning}.}
\label{fig:comparison}
\end{figure*}

Qualitative results obtained by our models on SALICON and MIT1003 validations sets, together with those of other state of the art models, are shown in Figure~\ref{fig:comparison}. As it can be noticed, our network is able to predict high saliency values on people, faces, objects and other predominant cues. It also produces good saliency maps when images do not contain strong saliency regions, such as when saliency is concentrated in the center of the scene or when images portray a landscape.
We notice, from a qualitative point of view, that the model can sometimes infer the relative importance of different people in the same scene, a human behaviour which saliency models still struggle to replicate, as discussed in~\cite{bylinskii2016should}.

\begin{table}[t]
\renewcommand{\arraystretch}{1.3}
\centering
\caption{Comparison results on MIT300 dataset~\cite{judd2012benchmark}. The results in bold indicate the best performing method on each evaluation metric. (*) indicates citations to non-peer reviewed texts. Methods are sorted by the NSS metric.}
\label{tab:mit300}
\begin{small}
\begin{tabular}{|l|cccccc|}
\hline
& \footnotesize{SIM} 	& \footnotesize{CC} 	& \footnotesize{sAUC} 	& \footnotesize{AUC}	& \footnotesize{NSS}	&  \footnotesize{EMD}	\\ \hline \hline
    \footnotesize{DSCLRCN~\cite{liu2016deep}} & \textbf{0.68} & \textbf{0.80} & 0.72 & 0.87 & \textbf{2.35} & 2.17 \\ \hline 
   
\textbf{\footnotesize{SAM-ResNet}} & \textbf{0.68}	& 0.78	& 0.70 & 0.87& 2.34 &  2.15 \\ \hline
\textbf{\footnotesize{SAM-VGG}} & 0.67	& 0.77	& 0.71	& 0.87	& 2.30	& 2.14  \\  \hline 
\footnotesize{DeepFix~\cite{kruthiventi2015deepfix}} & 0.67	& 0.78	& 0.71	& 0.87		& 2.26	& \textbf{2.04} \\ \hline 
\footnotesize{SALICON~\cite{huang2015salicon}}	& 0.60	& 0.74	& \textbf{0.74}	& 0.87	& 2.12	& 2.62	\\ \hline
\footnotesize{PDP~\cite{jetley2016end}} & 0.60	& 0.70	& 0.73	& 0.85	& 2.05	& 2.58   \\ \hline
\footnotesize{ML-Net~\cite{mlnet2016}} & 0.59	& 0.67	& 0.70	& 0.85	& 2.05	& 2.63 \\ \hline
\footnotesize{SalGAN~\cite{pan2017salgan} (*)} 	& 0.63	& 0.73	& 0.72	& 0.86	& 2.04	& 2.29	\\ \hline
\footnotesize{iSEEL~\cite{tavakoli2016exploiting}} & 0.57	& 0.65	& 0.68	& 0.84	& 1.78	& 2.72 \\ \hline
\footnotesize{SalNet~\cite{cPan}}	& 0.52		& 0.58		& 0.69	& 0.83		& 1.51	& 3.31	\\ \hline
\footnotesize{BMS~\cite{zhang2013saliency}}	& 0.51		& 0.55	& 0.65		& 0.83		& 1.41	& 3.35	\\ \hline
\footnotesize{Mr-CNN~\cite{liu2015predicting}}	& 0.48	& 0.48	& 0.69	& 0.79		& 1.37 	& 3.71	\\ \hline
\footnotesize{DeepGazeII~\cite{kummerer2016deepgaze}} & 0.46	 & 0.52	&  0.72	& \textbf{0.88}	& 1.29 	& 3.98	 \\ \hline
\footnotesize{GBVS~\cite{harel2006graph}}	& 0.48	& 0.48	& 0.63	& 0.81	& 1.24 	& 3.51	 \\ \hline
\footnotesize{eDN~\cite{vig2014large}}		& 0.41		& 0.45		& 0.62		& 0.82		& 1.14 	& 4.56	\\ \hline
\end{tabular}
\end{small}
\end{table}

\section{Conclusion}
We described a novel Saliency Attentive Model which can predict human eye fixations on natural images. The main novelty of the proposal is an Attentive Convolutional LSTM specifically designed to sequentially enhance saliency predictions. The same idea could potentially be employed in other tasks in which an image refinement is profitable. Furthermore, we captured an important property of human gazes by optimally combining multiple learned priors, and effectively addressed the downscaling effect of CNNs. The effectiveness of each component has been validated through extensive evaluation, and we showed that our model achieves state of the art results on two of the most important datasets for saliency prediction. Finally, we contribute to further research efforts by releasing the source code and pre-trained models of our architecture.

\begin{table}[t]
\renewcommand{\arraystretch}{1.3}
\centering
\caption{Comparison results on CAT2000 test set~\cite{CAT2000}. The results in bold indicate the best performing method on each evaluation metric.}
\label{tab:cat2000}
\begin{small}
\begin{tabular}{|l|cccccc|}
\hline
& \footnotesize{SIM} 	& \footnotesize{CC} 	& \footnotesize{sAUC} 	& \footnotesize{AUC}	& \footnotesize{NSS}	&  \footnotesize{EMD}\\ \hline \hline
\textbf{\footnotesize{SAM-ResNet}} & \textbf{0.77}	& \textbf{0.89}	& 0.58	& \textbf{0.88}	& \textbf{2.38}	& \textbf{1.04} \\ \hline   
\textbf{\footnotesize{SAM-VGG}} & 0.76	& \textbf{0.89} & 0.58	& \textbf{0.88}	& \textbf{2.38}	& 1.07 \\ \hline 
\footnotesize{DeepFix~\cite{kruthiventi2015deepfix}}	& 0.74	& 0.87	& 0.58	& 0.87	& 2.28	& 1.15	\\ \hline
\footnotesize{MixNet~\cite{dodge2017visual}}	& 0.66	& 0.76	& 0.58	& 0.86	& 1.92	& 1.63	\\ \hline 
\footnotesize{iSEEL~\cite{tavakoli2016exploiting}}	& 0.62	& 0.66	& \textbf{0.59}	& 0.84	& 1.67 & 1.78	\\ \hline 
\footnotesize{BMS~\cite{zhang2013saliency}}	& 0.61		& 0.67	& \textbf{0.59}	& 0.85	& 1.67	& 1.95	 \\ \hline
\footnotesize{eDN~\cite{vig2014large}}	& 0.52			& 0.54		& 0.55	& 0.85		& 1.30 	& 2.64	\\ \hline
\footnotesize{GBVS~\cite{harel2006graph}}		& 0.51	& 0.50	& 0.58	& 0.80		& 1.23 	& 2.99 \\ \hline
\end{tabular}
\end{small}
\end{table}

\section*{Acknowledgment}
We thank the organizers of the LSUN Challenge and of the MIT Saliency Benchmark for enabling us to compare with other published approaches.

This work was partially supported by JUMP project, funded by the Emilia-Romagna region within the POR-FESR 2014-2020 program. We acknowledge the CINECA award under the ISCRA initiative, for the availability of high performance computing resources and support. We also gratefully acknowledge the support of Facebook AI Research and NVIDIA Corporation with the donation of the GPUs used for this research.

\bibliographystyle{IEEEtran}
\bibliography{bibliography}

%


\begin{IEEEbiography}[{\includegraphics[width=1in,height=1.25in,clip,keepaspectratio]{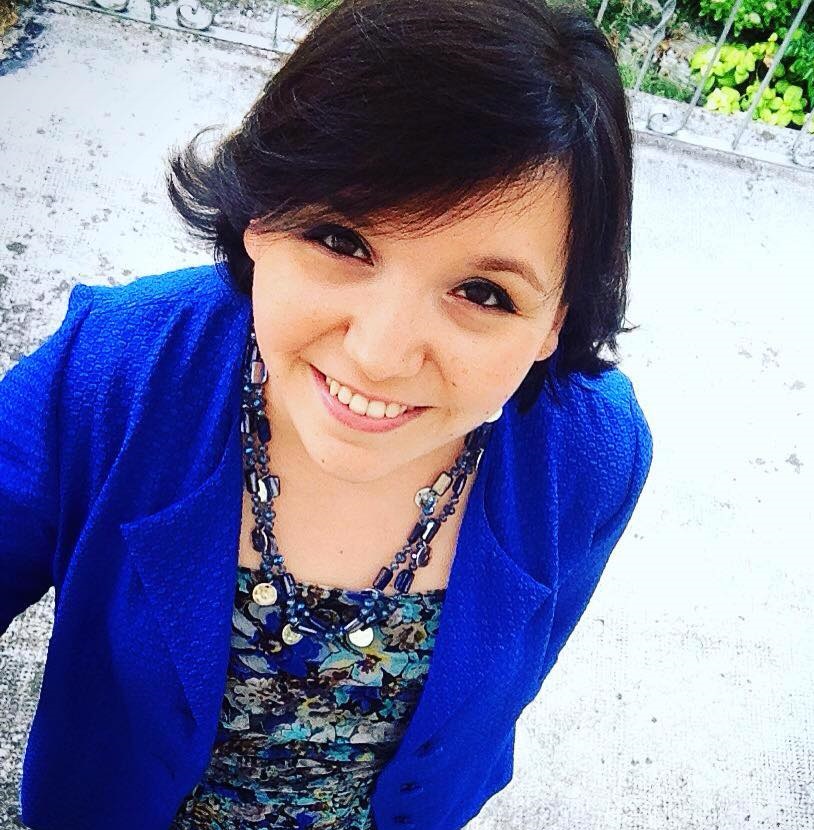}}]{Marcella Cornia} received the B.Sc. degree in Computer Science and the M.Sc. degree in Computer Engineering from the University of Modena and Reggio Emilia, Modena, Italy. She is currently pursuing the Ph.D. degree at the AImageLab Laboratory at the Department of Engineering ``Enzo Ferrari'' of the University of Modena and Reggio Emilia. Her research interests include visual saliency prediction, image captioning, and cross-modal retrieval. 
\end{IEEEbiography}

\begin{IEEEbiography}[{\includegraphics[width=1in,height=1.25in,clip,keepaspectratio]{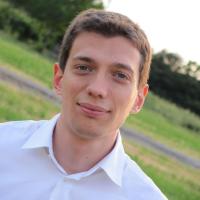}}]{Lorenzo Baraldi}
received the M.Sc. degree in Computer Engineering and the Ph.D. degree cum laude in Information and Communication Technologies from the University of Modena and Reggio Emilia, Modena, Italy, in 2014 and 2018, respectively. 
He is currently a Research Fellow with the Department of Engineering ``Enzo Ferrari'', University of Modena and Reggio Emilia. He was a Research Intern at Facebook AI Research (FAIR) in 2017. He has authored or coauthored more than 30 publications in scientific journals and international conference proceedings. His research interests include video understanding, deep learning and multimedia. He regularly serves as a Reviewer for international conferences and journals.
\end{IEEEbiography}


\begin{IEEEbiography}[{\includegraphics[width=1in,height=1.25in,clip,keepaspectratio]{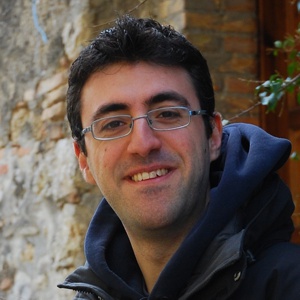}}]{Giuseppe Serra}
is currently an Assistant Professor with the University of Udine, Udine, Italy. He was a Visiting Scholar with Carnegie Mellon University, Pittsburgh, PA, USA, and at Telecom ParisTech/ENST, Paris, France, in 2006 and 2010, respectively. He has authored or coauthored more than 80 publications in scientific journals and international conference proceedings. His research interests include egocentric vision, and image and video analysis. Prof. Serra has been an Associate Editor for the IEEE Transactions on Human-Machine Systems since 2017. He was a Technical Program Committee member of several workshops and conferences. He regularly serves as a Reviewer for international conferences and journals such as CVPR and ACM Multimedia.
\end{IEEEbiography}

\begin{IEEEbiography}[{\includegraphics[width=1in,height=1.25in,clip,keepaspectratio]{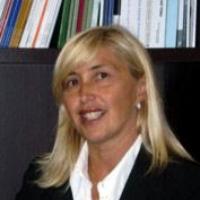}}]{Rita Cucchiara}
received the Ph.D. degree in Computer Engineering from the University of Bologna, Bologna, Italy, in 1989 and 1992, respectively. She is currently a Full Professor of computer engineering with the University of Modena and Reggio Emilia, Modena, Italy. She is the Director of the Research Center Softech-ICT, and heads the AImageLab Laboratory at the University of Modena and Reggio Emilia. She has authored or coauthored more than 350 papers in journals and international proceedings, and is a reviewer for several international journals. She has been a coordinator of several projects in computer vision and pattern recognition, and in particular on video surveillance, human behavior analysis, video understanding, human-centered searching in images and video, and big data for cultural heritage. Prof. Cucchiara is a Member of the ACM, of the IEEE Computer Society, and a Fellow of the IAPR. She is the President of the ``Associazione Italiana per la ricerca in Computer Vision, Pattern recognition e machine Learning'' (affiliated with IAPR), and a Member of the Advisory Board of the Computer Vision Foundation.
\end{IEEEbiography}




\end{document}